\documentclass[lettersize,journal]{IEEEtran}
\usepackage{amsmath,amsfonts}
\usepackage{algorithmic}
\usepackage{algorithm}
\usepackage{array}
\usepackage{textcomp}
\usepackage{stfloats}
\usepackage{url}
\usepackage{verbatim}
\usepackage{graphicx}
\usepackage{cite}

\usepackage{amssymb}
\usepackage{amsthm}
\newtheorem{theorem}{Theorem}

\newtheorem{definition}{Definition}
\newtheorem{proposition}{Proposition}

\usepackage{color}
\usepackage{subfigure}
\usepackage{graphicx}
\usepackage{multirow}
\usepackage{bm}
\usepackage{hyperref}
\hypersetup{colorlinks=true,
            linkcolor=cyan,
	    filecolor=blue,      
	    urlcolor=red,
	    citecolor=green}
\makeatletter

\newcommand{\Rmnum}[1]{\expandafter\@slowromancap\romannumeral #1@}
\makeatother
\begin{document}

\title{Improve Noise Tolerance of Robust Loss \\ via Noise-Awareness}

\author{Kehui Ding, Jun Shu, Deyu Meng, Zongben Xu
%~\IEEEmembership{Member,~IEEE,}
        % <-this % stops a space
\thanks{Kehui Ding, Jun Shu, Deyu Meng and Zongben Xu are with School of Mathematics and Statistics and Ministry of Education Key Lab of Intelligent Networks and Network Security, Xi,an Jiaotong University, Shaanxi, China.}% <-this % stops a space
\thanks{Jun Shu is the corresponding author: xjtushujun@gmail.com.}}

\markboth{Journal of \LaTeX\ Class Files,~Vol.~14, No.~8, August~2021}%
{Shell \MakeLowercase{\textit{et al.}}: A Sample Article Using IEEEtran.cls for IEEE Journals}


\maketitle

\begin{abstract}
Robust loss minimization is an important strategy for handling robust learning issue on noisy labels. Current approaches for designing robust losses involve the introduction of noise-robust factors, i.e., hyperparameters, to control the trade-off between noise robustness and learnability. However, finding suitable hyperparameters for different datasets with noisy labels is a challenging and time-consuming task. Moreover, existing robust loss methods usually assume that all training samples share common hyperparameters, which are independent of instances. This limits the ability of these methods to distinguish the individual noise properties of different samples and overlooks the varying contributions of diverse training samples in helping models understand underlying patterns. To address above issues, we propose to assemble robust loss with instance-dependent hyperparameters to improve their noise tolerance with theoretical guarantee. To achieve setting such instance-dependent hyperparameters for robust loss, we propose a meta-learning method which is capable of adaptively learning a hyperparameter prediction function, called Noise-Aware-Robust-Loss-Adjuster (NARL-Adjuster for brevity). Through mutual amelioration between hyperparameter prediction function and classifier parameters in our method, both of them can be simultaneously finely ameliorated and coordinated to attain solutions with good generalization capability. Four SOTA robust loss functions are attempted to be integrated with our algorithm, and comprehensive experiments substantiate the general availability and effectiveness of the proposed method in both its noise tolerance and performance. Meanwhile, the explicit parameterized structure makes the meta-learned prediction function ready to be transferrable and plug-and-play to unseen datasets with noisy labels. Specifically, we transfer our meta-learned NARL-Adjuster to unseen tasks, including several real noisy datasets, and achieve better performance compared with conventional hyperparameter tuning strategy, even with carefully tuned hyperparameters.
\end{abstract}


\section{Introduction}
\label{section1}

Deep neural networks have recently obtained remarkable performance on various applications \cite{ krizhevsky2012imagenet,he2016deep}. Its effective training, however, often requires to pre-collect large-scale finely annotated samples. When the training dataset contains certain amount of noisy (incorrect) labels, the overfitting issue tends to easily occur, naturally leading to their poor performance in generalization \cite{zhang2021understanding}. In fact, such biased training data are commonly encountered in practice, since the data are generally collected by coarse annotation sources, like crowdsourcing systems \cite{bi2014learning} or search engines \cite{liang2016learning,zhuang2017attend}. Such robust deep learning issue has thus attracted much attention recently in the machine learning field \cite{frenay2013classification,song2022learning}.

One of the most classical methods for handling this issue is to employ robust losses, which are not unduly affected by noisy labels, to replace the conventional noise-sensitive ones to guide the training process \cite{manwani2013noise}. For example, as compared with the commonly used cross entropy (CE) loss which has been proved to be not generally robust to label noise, the mean absolute error (MAE), as well as the simplest 0-1 loss for classification, can be more robust against noisy labels \cite{ghosh2017robust} due to their evident suppression to large loss values, and thus inclines to reduce the negative influence brought by the outlier samples with evidently corrupted labels. Beyond other robust learning techniques against noisy labels, like sample selection \cite{han2018co,pleiss2020identifying,wu2020topological,mirzasoleiman2020coresets,kim2021fine}, sample reweighting \cite{kumar2010self,chang2017active,wang2017robust,jiang2018mentornet,ren2018learning,shu2019meta,zhao2021probabilistic}, loss correction \cite{sukhbaatar2014training,goldberger2016training,patrini2017making,hendrycks2018using,shu2020meta}, and label correction \cite{tanaka2018joint,arazo2019unsupervised,li2020dividemix,liu2020early,cheng2020learning,wu2021learning,shu2022cmw}, such robust-loss-designing methodology is superior in its concise implementation scheme and solid theoretical basis of robust statistics and generalization theory \cite{huber2004robust,masnadi2008design,liu2014robust}.
%\IEEEpubidadjcol

Manwani et al.\cite{manwani2013noise} formally defines a “noise-tolerant” loss if the minimization under it with noisy labels would achieve the same solution as that with noise-free labels. Along this research line, besides the aforementioned MAE and 0-1 loss, multiple forms of robust losses have been studied against such robust learning issue on noisy labels. For example, 0-1 loss is verified to be robust for binary classification problem \cite{manwani2013noise,ghosh2015making}. However, since 0-1 loss is not continuous and the corresponding learning algorithm is hardly to be efficiently and accurately executed, many surrogate loss functions have been proposed to approximate it \cite{bartlett2006convexity,masnadi2008design,nock2008efficient}, such as ramp loss \cite{brooks2011support} and unhinged loss \cite{van2015learning}, which are also proved to be robust to label noise under certain conditions \cite{ghosh2015making}. In addition, \cite{ghosh2015making} introduces a sufficient condition, called symmetric loss condition, for judging whether the loss function is noise-tolerant. \cite{ghosh2017robust} further extends the definition with the symmetric loss condition to multi-classification. Although conventional losses satisfying the symmetric loss condition evidently alleviate the overfitting problem on datasets with noisy labels, it has been found empirically that they usually encounter the problem of insufficient learning on large datasets \cite{zhang2018generalized,wang2019symmetric,ma2020normalized}. Inspired by the above problem, several losses have further been designed recently by introducing some noise-robust factors to modify CE loss and proved to be robust against noise and guarantee a more sufficient learning on all training samples, like the GCE \cite{zhang2018generalized}, SL \cite{wang2019symmetric}, PolySoft \cite{gong2018decomposition} and JS \cite{englesson2021generalized}, etc.

Although these robust loss functions are helpful in improving the learning performance against noisy labels, they still have evident deficiencies in practice. On the one hand, they inevitably involve noise-robust factor, i.e., hyperparameter(s), to control their robustness extents against noisy labels. These hyperparameters need to be manually preset or tuned by heuristic strategies like cross-validation, naturally conducting efficiency issue and difficulty in practical implementations. On the other hand, label noise problem is notoriously complicated since noisy labels have different noise structure forms, such as \textit{noise rate, noise type} and \textit{classes property}, and such noise difference exists over different samples. Current robust loss methods, however, are not noise-aware \footnote[1]{The robust loss is noise-aware, implying that the loss of each sample relies on its individual noise information.}, and introduce a consistent hyperparameter(s) bias across all training samples, which possibly leads to the issue that the learned classifier barely reaches the optimal performance, especially when we know insufficient knowledge of underlying data or the label conditions are too complicated.

\begin{figure*}[t]
  \centering
  \includegraphics[width=0.95\textwidth]{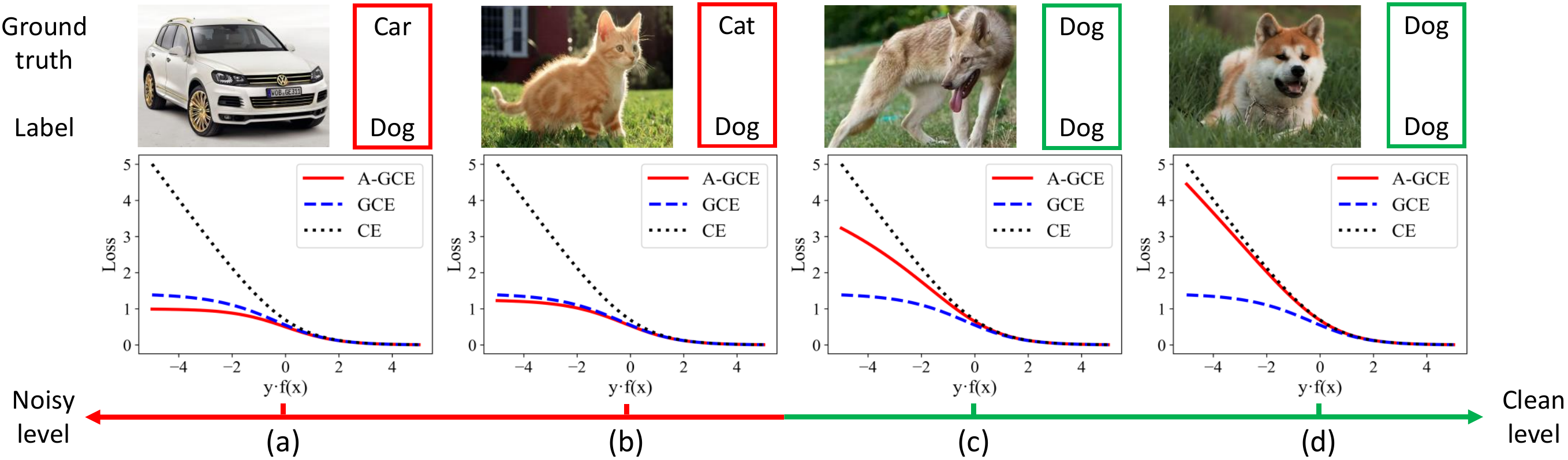}
  \caption{Illustration of the employed loss functions during training for different samples, including CE, GCE \cite{zhang2018generalized}, and noise-aware GCE (denoted by A-GCE), where (a-b) are noisy samples and (c-d) are clean samples. It is evident that with the individual noise properties distinguishing, A-GCE shows adaptive robust loss setting ability for finely reflecting such noise variance.}
  \label{figure1}
  \vspace{-0.5cm}
\end{figure*}

To alleviate these issues, two problems need to be intrinsically considered. 1) \textbf{How to improve the noise-aware ability of current robust loss methods to adapt to the complicated noise structures?} As depicted by the curves of CE (black) and GCE (blue) in Fig.\ref{figure1}(a-b), compared to CE loss overfitting the noisy samples, current robust loss methods mitigate the impact of these samples. This is achieved by assigning them smaller loss values through the introduction of specific noise-robust factors (blue curves are lower than black curves). Consequently, the training process becomes more robust against noisy labels. However, the noise-robust hyperparameters of robust losses are instance-independent and shared across all training samples, i.e., the blue curves representing robust losses of noisy samples in Fig.\ref{figure1}(a-b) are the same as those of clean samples in Fig.\ref{figure1}(c-d). While attempting to suppress the negative effects of noisy samples, they also tend to inadvertently diminish the positive effects of clean samples (as illustrated in Fig.\ref{figure1}(c-d), for clean samples, blue curves are significantly lower than curves of CE which possess stronger learnability \cite{zhang2018generalized,englesson2021generalized}). Clearly, the employment of such simple instance-independent hyperparameter settings exhibits limitations in adapting to diverse and complicated noise structures.

One natural idea involves assembling robust loss with instance-dependent hyperparameters, as this enables the learning of appropriate hyperparameters for samples with distinct characteristics to control their influence during training. More precisely, to further weaken the adverse impact of noisy samples during training, our expectation is that the loss values of noisy samples would be reduced. As shown in Fig.\ref{figure1}(c-d), the red curves denoting noise-aware GCE impose noisy samples on smaller loss values than blue curves (GCE). In order to enable the model to better discern underlying patterns, we should amplify the beneficial impact of clean samples more. In the case shown in Fig.\ref{figure1}(a-b), the noise-aware losses of clean samples are larger than GCE losses of clean samples.

2) \textbf{How to learn instance-dependent hyperparameter configuration for robust loss?} One strategy is to directly learn the sample-wise hyperparameters for robust loss. However, its computational complexity scales with the dataset size, making it highly inefficient and infeasible especially for larger datasets. Additionally, we theoretically find that the noise-aware ability of sample-wise hyperparameters relies on the noise properties of each sample (conclusions of Pro.\ref{pro1}). The relationship between instance-dependent hyperparameters learned directly and noise properties of training samples is implicit, which means this relationship is difficult to constrain and lacks interpretability. Therefore, we propose to use an explicit hyperparameter prediction function to learn such instance-dependent hyperparameters conditioned on the information of each sample by leveraging current meta learning techniques \cite{finn2017model,hospedales2021meta,JMLR:v24:21-0742,shu2022cmw,shu2022mlr}. It is expected to produce different hyperparameters for noisy and clean samples via the prediction function, and then naturally to improve the noise-aware ability of robust loss. We call this prediction function as Noise-Aware-Robust-Loss-Adjuster (NARL-Adjuster for brevity). More precisely, we employ MLPs to parameterize the NARL-Adjuster. It is known that MLP network is with strong approximation ability \cite{hornik1989multilayer}, and thus the proposed NARL-Adjuster is potentially useful to learn a noise-aware hyperparameter prediction function to adapt to complicated noise structures.

In a nutshell, this paper mainly makes the following contributions.
\begin{itemize}
\item{We find that current robust loss methods have limited flexibility to adapt to the complicated noise structures, which are attributed to their adopted strategy of shared instance-independent hyperparameters, i.e., a unique hyperparameter presetting over all samples. Differently, we propose to assemble robust loss with instance-dependent hyperparameters, i.e., adaptive hyperparameter settings on different samples based on their specific noise property (as shown in Fig.\ref{figure1}), to improve their noise tolerance of addressing various noise structures.}	
\item{To learn instance-dependent hyperparameters for robust loss, we propose to use a hyperparameter prediction function to adaptively output sample-wise hyperparameters. We learn the prediction function in a meta-learning manner, and the algorithm realizes a mutual amelioration between sample-wise hyperparameters prediction involved in the robust loss and classifier network learning.}	
\item{The proposed method is model-agnostic, and is substantiated to be valid in improving the noise-aware ability of four kinds of SOTA robust loss methods, including GCE\cite{zhang2018generalized}, SL\cite{wang2019symmetric},  PolySoft\cite{gong2018decomposition} and JS\cite{englesson2021generalized}. Comprehensive experiments on different noise structures, like symmetric, asymmetric and instance-dependent label noises, as well as variations in noise rates and the number of classes, demonstrate the noise-tolerance superiority of the proposed method beyond conventional robust loss approaches.}
\item{We theoretically prove that the proposed noise-aware robust loss has a strong noise-tolerance property under certain conditions, which implies that our method is with a sound theoretical guarantee and potential usefulness.}	
\item{We further study the transferability of NARL-Adjuster. Different from current robust loss methods requiring to pre-specify an ad-hoc hyperparameter tuning scheme, the learned hyperparameter prediction function is verified to be readily used in a plug-and-play manner, and can be directly deployed on unseen datasets to produce proper noise-aware hyperparameters, without extra hyperparameters required to be tuned.}
\end{itemize}

The paper is organized as follows. Section \ref{section2} reviews the related works. Necessary Notations, definitions and theories of robust loss used in this paper are presented in Section \ref{section3}. Section \ref{section4} introduces our main algorithm for noise-aware robust loss learning. Experiments are demonstrated in Section \ref{section5}, and a conclusion is finally made.

\section{Related Work}
\label{section2}

\subsection{Deep learning with noisy labels}

There are various approaches raised for handling the robust learning issue under noisy labels, which can be roughly divided into five categories: sample selection, loss correction, label correction, sample reweighting and robust loss methods. We firstly introduce the former four approaches.

The sample selection approach aims to find clean samples to train neural networks efficiently and design filtering rules by small loss \cite{han2018co}, logits vector \cite{pleiss2020identifying}, topological information \cite{wu2020topological,kim2021fine} and Jacobian matrix \cite{mirzasoleiman2020coresets}. Comparatively, the loss correction approach assumes that there exists a noise transition matrix defining the probability of one class changed to another. Typically, \cite{sukhbaatar2014training,goldberger2016training} model the matrix using a linear layer on the top of the DNNs. Besides, Forward \cite{patrini2017making} uses the noise transition matrix to modify the loss function. GLC \cite{hendrycks2018using} and Meta-Transition \cite{shu2020meta} utilize additional meta data to estimate the transition matrix. Comparatively, label correction approach tries to correct noisy labels to their true ones via a supplemental clean label inference step \cite{tanaka2018joint,arazo2019unsupervised,li2020dividemix,liu2020early,cheng2020learning,wu2021learning,shu2022cmw}. Besides, the main idea of sample reweighting approach is to assign weights to losses of all training samples, and iteratively update these weights based on the loss values during the training process \cite{kumar2010self,jiang2014easy,zhao2015self}. Such a loss-weight function is generally set as monotonically decreasing, enforcing a learning effect that samples with larger loss, more possible to be noisy labeled as compared with small loss samples, with smaller weights to suppress their effects to training. In this manner, the negative influence of noisy labels can be possibly alleviated. Interestingly, it can always be proved that this monotonically decreasing weighting function makes this re-weighting learning process equivalent to solving an implicit robust loss function \cite{meng2017theoretical}, which constructs a close relationship between this strategy with robust loss approach. Recently, some advanced sample reweighting methods have been raised inspired by the idea of meta-learning \cite{shu2019meta, zhao2021probabilistic,shu2022cmw}, which possesses a data-driven weighting scheme to the conventional hand-designed reweighting strategies. This makes them able to deal with more general data bias cases other than noisy labels, like class imbalance.

\subsection{Robust loss approach}

Based on the noise-tolerant definition given by \cite{manwani2013noise}, it has been proved that 0-1 loss, sigmoid loss, ramp loss, and probit loss are all noise-tolerant under some mild conditions \cite{ghosh2015making}. For any surrogate loss function in binary classification, \cite{natarajan2013learning} proposes an approach to help them to be noise-tolerant. Recently, \cite{lyu2019curriculum} proposes curriculum loss,  a tighter upper bound of the 0-1 loss, which can adaptively select samples for training as a curriculum learning process. Generalized to multi-class problem, Mean Absolute Error (MAE) is proved to be robust against symmetric and class-conditional label noise \cite{ghosh2017robust}.

Under the advanced noise-robust understanding provided by \cite{ghosh2017robust}, some new robust losses have been designed on the basis of classical CE loss very recently, expected to be finely performable in real practice. \cite{zhang2018generalized} demonstrates that it is hard to train DNN with MAE, and proposes to combine MAE and CE losses to obtain a new loss function, GCE, which behaves very like a weighted MAE, to handle the noisy label issue. \cite{wang2019symmetric} observes that the learning procedure of DNNs with CE loss is class-biased, and proposes a Reverse Cross Entropy (RCE) to further enhance robust learning. \cite{feng2020can} applies Taylor Series to derive a surrogate representation of CE, which can also relate MAE and CE by varying the order of Taylor Series. Besides, \cite{ma2020normalized} theoretically verifies that a normalization can help losses to be robust to noisy labels, and proposes active passive loss associating two kinds of robust loss namely active loss and passive loss to overcome the insufficient learning issue. \cite{englesson2021generalized} introduces Jensen-Shannon divergence as a loss function and generalizes the proposed loss function to multiple distributions. Though the loss has been proved to be noise-tolerant to asymmetric and symmetric noisy labels under some conditions, it uses shared instance-independent hyperparameters with a permanent hyperparameter(s) bias. This possibly leads to the issue that the learned classifier barely reaches the optimal performance, especially when we know insufficient knowledge underlying data or the label conditions are too complicated. In this paper, we propose to assemble robust loss with instance-dependent hyperparameters, which is expected to improve their noise-tolerance to adapt to the complicated noise structures.

\subsection{Methods on learning adaptive loss}

Some other methods have also been proposed attempting to directly learn a good proxy for an underlying evaluation loss. For example, learning to teach \cite{wu2018learning} dynamically learns the loss through a teacher network outputting the coefficients matrix of the general loss function. \cite{xu2018autoloss} learns a discrete optimization schedule that alternates between different loss functions at different time-points. Adaptive Loss Alignment \cite{huang2019addressing} extends work in \cite{wu2018learning} to loss-metric mismatch problem. \cite{grabocka2019learning} tries to learn the surrogate losses for non-differentiable and non-decomposable loss function. \cite{Li_2019_ICCV} designs a search space containing almost all popular loss functions and  dynamically optimizes sampling distribution of loss functions. \cite{gao2021searching} chooses Taylor series approximations of functions as the searching space, and the learned loss functions with fixed hyperparameters can be directly employed to new datasets with noisy labels. Building upon MAML \cite{finn2017model}, \cite{baik2021meta} adaptively meta-learns a loss function that adapts to each few-shot learning task. Though these methods have achieved automatically learning loss for various tasks, their attained losses generally have relatively complicated forms and with little theoretical guarantee. Comparatively, our method is built on the robust loss with relatively more solid theoretical foundation, better explainability for the learned loss, simpler realization form and easier transferability across different datasets.

\section{Noise-Aware Robust Loss Method}
\label{section3}

\subsection{Preliminaries and Notations}
\label{section31}

We consider the problem of $c$-class classification. Let $\mathcal{X}\subset\mathbb{R}^d$ be the feature space, and $\mathcal{Y}=\{1,2,\cdots,c\}$ be the label space. Assuming that the DNN architecture is with a softmax output layer. Denoting the network as a function with input $\mathbf{x}\in\mathcal{X}$ and output as $f(\mathbf{x};\mathbf{w})$, where $f:\mathcal{X}\to\mathbb{R}^c$, and $\mathbf{w}$ represents the network parameters. ${f}_{j}(\mathbf{x};\mathbf{w})$, representing the $j$-th component ($j=\{1,\cdots,c\}$) of $f(\mathbf{x};\mathbf{w})$, and thus satisfies $\sum_{j=1}^{c}{f}_{j}(\mathbf{x};\mathbf{w})=1$, ${f}_{j}(\mathbf{x};\mathbf{w})\geq0$. Given training data $D=\{(\mathbf{x}_{i},\mathbf{y}_{i})\}_{i=1}^{N} \in (\mathcal{X}\times \mathcal{Y})^{N}$, for any loss function $\mathcal{L}$, the (expection) risk of the network classifier is defined as $R_{\mathcal{L}}(f)=E_{D}[\mathcal{L}(f(\mathbf{x},\mathbf{w}),\mathbf{y})]$. The commonly used CE loss can be written as:
\begin{equation}\label{eq1}
  \mathcal{L}_{CE}(f(\mathbf{x}_{i};\mathbf{w}),\mathbf{y}_{i})=-\sum_{j=1}^{c} y_{ij}\log f_{j}(\mathbf{x}_i;\mathbf{w}),
\end{equation}
where $y_{ij}$ denotes the $j$-th component of $\mathbf{y}_{i}$. Generally in all components of $\mathbf{y}_{i}$, only one is 1 and all others are 0.

As for training data with noisy labels $\hat{D}$, we denote the noisy label of $\mathbf{x}$ as $\hat{\mathbf{y}}$, in contrast to its clean label $\mathbf{y}$, and the dataset can be divided into clean part ($\hat{\mathbf{y}}=\mathbf{y}$) and noisy part ($\hat{\mathbf{y}}\ne\mathbf{y}$). The robustness theories of conventional robust loss functions are all based on assumptions that label noise is independent with instances, so we define the probability that the sample belongs to noisy part as
\begin{equation}\nonumber
  \begin{aligned}
  \eta&=\sum_{j,\mathbf{e}_j \ne \mathbf{y}_{\mathbf{x}}}\eta_{\mathbf{y}_{\mathbf{x}}j}=\sum_{j,\mathbf{e}_{j} \ne \mathbf{y}_{\mathbf{x}}}p(\hat{\mathbf{y}}=\mathbf{e}_{j}|\mathbf{y}=\mathbf{y}_{\mathbf{x}},\mathbf{x})\\
  &=\sum_{j,\mathbf{e}_{j} \ne \mathbf{y}_{\mathbf{x}}}p(\hat{\mathbf{y}}=\mathbf{e}_{j}|\mathbf{y}=\mathbf{y}_{\mathbf{x}}),     
  \end{aligned}
\end{equation}
where $\mathbf{e}_{j}$ is a $c$-dimensional vector with one at index $j$ and zero elsewhere. If $\eta_{\mathbf{y}_{\mathbf{x}}j}=\frac{\eta}{c-1}$ for all $j$ satisfies $\mathbf{e}_j\ne \mathbf{y}_{\mathbf{x}}$, the noise is termed as uniform or symmetric, and noise is called asymmetric or class dependent if the dependence of $\eta_{\mathbf{y}_{\mathbf{x}}j}$ on $\mathbf{x}$ is only through $\mathbf{y}_{\mathbf{x}}$. Correspondingly, for loss function $\mathcal{L}$, the risk of the classifier under label noise rate $\eta$ is denoted as $R^{\eta}_{\mathcal{L}}(f)=E_{\hat{D}}[\mathcal{L}(f(\mathbf{x},\mathbf{w}),\hat{\mathbf{y}})]$. \cite{manwani2013noise} provides the definition of noise-tolerant loss function as follows.

\begin{definition}\label{def1}
  A loss function $\mathcal{L}$ is called noise-tolerant if $f^{*}$ is also the global minimizer of $R_{\mathcal{L}}^{\eta}(f)$, where $f^{*}$ is the global minimizer of $R_{\mathcal{L}}(f).$
\end{definition}

\begin{definition}\label{def2}
  If a loss function $\mathcal{L}$ satisfies that, for some constant $C$, $$\sum_{j=1}^{c}\mathcal{L}(f(\mathbf{x};\mathbf{w}),j)=C, \forall \mathbf{x}\in\mathcal{X}, \forall f \in \mathcal{F},$$
  we call the loss function symmetric. 
\end{definition}

Ghosh, et al. \cite{ghosh2017robust} proves that if a multi-classification loss function is symmetric, then this loss is noise-tolerant under some mild assumptions. To make the above condition more feasible in practice, \cite{zhang2018generalized} further relaxes this definition as the bounded loss condition and proves that the loss with such new condition also possesses certain robustness property.

\begin{definition}\label{def3}
  If a loss function $\mathcal{L}$ satisfies that, for some constant $C_{L}$ and $C_{U}$, $$0\leq C_{L}\le\sum_{j=1}^{c}\mathcal{L}(f(\mathbf{x};\mathbf{w}),j)\le C_{U}, \forall \mathbf{x}\in\mathcal{X}, \forall f,$$
  we call the loss function bounded. 
\end{definition}

\subsection{Typical Robust Loss Methods}
\label{section32}

Here, we introduce several typical robust loss methods, as well as their theorical guarantee of the robustness.

\textbf{Generalized Cross Entropy (GCE).} To exploit the benefits of both the noise-tolerant property of MAE and the implicit weighting scheme of CE for better learning, Zhang et al. \cite{zhang2018generalized} proposed the GCE loss as follows:
\begin{equation}\label{eq2}
    \mathcal{L}_{GCE}(f(\mathbf{x}_{i};\mathbf{w}),\mathbf{y}_{i};{\color{red}q})=\frac{(1-f_{j_{i}}(\mathbf{x}_i)^{\color{red}q})}{\color{red}q},
\end{equation}
where $j_i$ denotes the $j$'s index if the term $y_{ij}=1$ for each $i$, and $q\in{(0, 1]}$. GCE loss is generated to the CE when $q$ approaches to 0 and becomes MAE loss when $q=1$.

\textbf{Symmetric Cross Entropy (SL).} Wang et al. \cite{wang2019symmetric} proposed the Reverse Cross Entropy (RCE) robust loss :$$\mathcal{L}_{RCE}(f(\mathbf{x}_{i};\mathbf{w}),\mathbf{y}_{i})=-\sum_{j\ne j_i}Af_j(\mathbf{x}_i;\mathbf{w}),$$ where $A<0$ is a preset constant. They further combine it with CE loss to obtain SL loss defined as follows:
\begin{equation}\label{eq3}
    \mathcal{L}_{SL}(f(\mathbf{x}_{i};\mathbf{w}),\mathbf{y}_{i};{\color{red}\gamma_1},{\color{red}\gamma_2})={\color{red}\gamma_1}\mathcal{L}_{CE}+{\color{red}\gamma_2}\mathcal{L}_{RCE},
\end{equation}
where $\gamma_1,\gamma_2>0$ are the combining coefficients.

\textbf{Polynomial Soft Weight Loss (PolySoft).} Self-paced learning (SPL) is a typical sample reweighting strategy to handle noisy labels by setting monotonically decreasing weighting function \cite{kumar2010self,liu2014robust,zhao2015self}. It has been proved that such re-weighting learning process is equivalent to solving an latent robust loss \cite{meng2017theoretical}, and it thus can also be categorized into the robust loss method. Recently, Gong et al. \cite{gong2018decomposition} proposed a polynomial soft weighting scheme for SPL, which can generally approximate monotonically decreasing weighting functions. By setting the CE loss as the basis loss form, the latent robust loss of this method is:
\begin{equation}\label{eq4}
  \begin{aligned}
    &\mathcal{L}_{Poly}(f(\mathbf{x}_{i};\mathbf{w}),\mathbf{y}_{i};{\color{red}\lambda},{\color{red}d})=\\
    &\begin{cases}
      \frac{({\color{red}d}-1){\color{red}\lambda}}{{\color{red}d}}[1-(1-\frac{\mathcal{L}_{CE}}{{\color{red}\lambda}})^{\frac{{\color{red}d}}{{\color{red}d}-1}}], &\mathcal{L}_{CE}<{\color{red}\lambda}\\
      \frac{({\color{red}d}-1){\color{red}\lambda}}{{\color{red}d}}, &\mathcal{L}_{CE}\ge{\color{red}\lambda}
    \end{cases}    
  \end{aligned}
\end{equation}
where $\mathcal{L}_{CE}$ is defined as in Eq. (\ref{eq1}).

\textbf{Jensen-Shannon divergence loss (JS).} Englesson et al., \cite{englesson2021generalized} employed the following JS divergence to deal with noisy labels:
\begin{equation}\label{eq5}
  \begin{aligned}
    &\mathcal{L}_{JS}(f(\mathbf{x}_{i};\mathbf{w}),\mathbf{y}_{i};{\color{red}\pi_1},{\color{red}\pi_2})=\\
    &\frac{{\color{red}\pi_1}D_{KL}(\mathbf{y}_{i}||\mathbf{m})+{\color{red}\pi_2}D_{KL}(f(\mathbf{x}_{i};\mathbf{w})||\mathbf{m})}{Z},
  \end{aligned}
\end{equation}
where $\mathbf{m}={\color{red}\pi_1}\mathbf{y}_{i}+{\color{red}\pi_2}f(\mathbf{x}_{i};\mathbf{w})$ and $Z=-(1-{\color{red}\pi_1})\log(1-{\color{red}\pi_1})$. Similar to GCE, JS loss can also be seen as a generalized mixture of CE and MAE via introducing the constant factor $Z$ (When $\pi_1 \rightarrow 0$, $\mathcal{L}_{JS} \rightarrow \mathcal{L}_{CE}$ and when $\pi_1 \rightarrow 1$, $\mathcal{L}_{JS} \rightarrow \frac{1}{2}\mathcal{L}_{MAE}$).

We then focus on the robustness theories of conventional robust loss functions. It has been proved that GCE \cite{zhang2018generalized} and JS \cite{englesson2021generalized} are bounded and RCE \cite{wang2019symmetric} exhibits symmetric property in the original papers. Thus, the above three loss functions possess theoretical robustness properties under symmetric label noise and asymmetric label noise. We can further prove that the PolySoft loss \cite{gong2018decomposition} also possesses these properties, as provided in the following theorem.

\begin{theorem}\label{thm1}
  Under the symmetric noise with $\eta \le 1 - \frac{1}{c}$, $f^*$ is the global minimizer of $R_{\mathcal{L}}(f)$ and $\hat{f}$ is the global minimizer of $R^{\eta}_{\mathcal{L}}(f)$, the following results then hold: $0 \le R^{\eta}_{\mathcal{L}}(f^{*}) - R^{\eta}_{\mathcal{L}}(\hat{f}) \le \frac{\eta(C_U-C_L)}{c-1}$, and $\frac{-\eta(C_U-C_L)}{c-1-\eta c} \le R_{\mathcal{L}}(f^{*}) - R_{\mathcal{L}}(\hat{f}) \le 0$, in which:\\
  (1) for the GCE loss in Eq.(\ref{eq2}), $C_L=\frac{c-c^{1-q}}{q}$ and $C_U=\frac{c-1}{q}$;\\
  (2) for the RCE loss in Eq.(\ref{eq3}), $C_L=C_U$;\\
  (3) for the PolySoft loss in Eq.(\ref{eq4}), $C_L=\frac{c(d-1)logc}{d}$ and $C_U=\frac{c(d-1)\lambda}{d}$;\\
  (4) for the JS loss in Eq.(\ref{eq5}), $C_L=\sum_{i=1}^{c}\mathcal{L}(\mathbf{e}_{i},\mathbf{u})$ and $C_U=\sum_{i=1}^{c}\mathcal{L}(\mathbf{e}_{i},\mathbf{e}_{1})$, where $\mathbf{u}$ is the uniform distribution.
\end{theorem}

The aforementioned theorem asserts that, under symmetric label noise, $R^{\eta}_{\mathcal{L}}(f^{*}) - R^{\eta}_{\mathcal{L}}(\hat{f})$ of four mentioned robust loss functions are bounded. Furthermore, when the upper bound is equal to zero, it signifies that the corresponding loss function exhibits noise-tolerant characteristic (Def.\ref{def1}). The theorem about asymmetric label noise is listed in supplementary material Section \Rmnum{1}.

\subsection{Limitations of Current Robust Loss Methods}
\label{section33}

It can be observed that all aforementioned loss functions contain hyperparameter(s), e.g., $q$ in $\mathcal{L}_{GCE}$ (Eq.(\ref{eq2})), $\gamma_1$, $\gamma_2$ in $\mathcal{L}_{SL}$ (Eq.(\ref{eq3})), $\lambda$, $d$ in $\mathcal{L}_{Poly}$ (Eq.(\ref{eq4})) and $\pi_1$, $\pi_2$ in $\mathcal{L}_{Poly}$ (Eq.(\ref{eq5})). For simplicity, we use $\mathcal{L}_{Robust}(f(\mathbf{x}_{i}; \mathbf{w}),\mathbf{y}_{i};\Lambda)$ to represent robust loss, where $\Lambda$ is the hyperparameter(s) of the robust loss function. Thus the classification network parameters are trained by optimizing the following empirical risk minimization problem under certain robust loss function:
\begin{equation}\label{eq6}
  \mathbf{w}^{*}=\arg\min_{\mathbf{w}}\frac{1}{N}\sum_{i=1}^{N}\mathcal{L}_{Robust}(f(\mathbf{x}_{i}; \mathbf{w}),\mathbf{y}_{i};\Lambda).
\end{equation}

These hyperparameters $\Lambda$ have specific physical meanings in combating noisy labels. Since the physical meanings of most robust losses are related to CE and MAE, we first analyze the above two losses from the perspective of gradients:
\begin{equation}\nonumber
  \begin{aligned}
   &\frac{\partial \mathcal{L}_{CE}(f(\mathbf{x}_{i};\mathbf{w}),\mathbf{y}_{i})}{\partial \mathbf{w}}=-\frac{1}{f_{\mathbf{y}_{i}}(\mathbf{x}_{i};\mathbf{w})}\nabla_{\mathbf{w}}f_{\mathbf{y}_{i}}(\mathbf{x}_{i};\mathbf{w}),\\
   &\frac{\partial \mathcal{L}_{MAE}(f(\mathbf{x}_{i};\mathbf{w}),\mathbf{y}_{i})}{\partial \mathbf{w}}=-\nabla_{\mathbf{w}}f_{\mathbf{y}_{i}}(\mathbf{x}_{i};\mathbf{w}).\end{aligned}
\end{equation}
It can be observed that the gradient of CE contains an additional weighted term $\frac{1}{f_{\mathbf{y}_{i}}(\mathbf{x}_{i};\mathbf{w})}$, which is beneficial to effectively learn samples associated with lower prediction probabilities. Nevertheless, the incorporation of the weighted term gives rise to an overfitting issue concerning label noise, as it tends to treat the majority of simple noisy instances, which are classified with relatively high confidence into classes that do not match the labels, as hard instances. The MAE, treating every sample equally, mitigates the impact of noisy labels.

GCE (JS) with the hyperparameter $q = 1$ ($\pi_1 \rightarrow 1$) exhibits stronger robustness and demonstrates an enhanced capacity to alleviate the deleterious effects of noisy labels (similar mechanism with MAE loss); in contrast, GCE (JS) with the hyperparameter $q \rightarrow 0$ ($\pi_1 \rightarrow 0$) exhibits comparatively weaker robustness, while it facilitates more learnability (similar mechanism with CE loss). Additionally, the hyperparameters $\gamma_1, \gamma_2$ of SL can be regarded as the linear combination coefficients of CE and MAE. Thus, the noise-robust factors in the previous three losses serve to govern the trade-off between CE and MAE, striking a balance between the robustness of loss and its capacity for effective learning.

Regarding PloySoft loss, the hyperparameter $\lambda$ can be interpreted as an indicator of noisy labels (age parameter in self-paced learning \cite{kumar2010self}), wherein samples with CE losses greater than $\lambda$ are more likely to correspond to noisy labels, while samples with CE losses less than $\lambda$ are more likely to correspond to clean labels.

In the context of handling diverse label noise, it is feasible to enhance model training efficacy by adjusting the hyperparameters of the loss function. However, the task of selecting a suitable hyperparameter configuration often presents a challenging endeavor. For instance, hyperparameters that offer superior robustness to label noise may simultaneously hinder the model's capacity to capture informative patterns in the data, leading to a compromise in the overall predictive performance of the model. Therefore, for every classification task with noisy labels, it is necessary to invest time in searching for appropriate hyperparameters.

Moreover, it is unreasonable to assume that all training samples have the same hyperparameters against noisy labels. Firstly, this is due to the fact that while they tend to suppress the negative effects of noisy samples, they also unexpectedly incline to weaken the positive effects of clean samples. Secondly, employing the consistent hyperparameter configuration for all samples neglects the inherent characteristics of individual instances and their relative importance to the learning process.   Each sample plays a distinct role in shaping the model's comprehension of underlying patterns, and a uniform setting strategy dismisses this critical aspect. Hence, the consistent hyperparameter(s) bias may potentially impede the improvement of model performance.

\subsection{Improving Noise Tolerance of Robust Loss via Noise-Awareness}
\label{section34}

To tackle the above issues, one natural idea is to assemble robust loss with instance-dependent hyperparameters, which aims to properly assign different hyperparameters for samples with various noise information, i.e., noise-awareness. Specifically, we re-write the robust loss function in Eq.(\ref{eq6}) as:
\begin{equation}\label{eq7}	
  \mathbf{w}^{*}=\arg\min_{\mathbf{w}}\frac{1}{N}\sum_{i=1}^{N}\mathcal{L}_{Robust}(f(\mathbf{x}_{i}; \mathbf{w}),\mathbf{y}_{i};\Lambda(\mathbf{x}_{i},\mathbf{y}_{i} )),
\end{equation}
where the loss for each sample is equipped with the instance-dependent hyperparameter(s) $\Lambda(\mathbf{x}_{i},\mathbf{y}_{i} )$. In accordance with the distinctive physical significance of hyperparameters, as elaborated in Section \ref{section33}, it is anticipated to adaptively set appropriate hyperparameters on noisy samples to more suppress their negative impact during training, while on clean samples to more amplify their positive effects throughout the learning process.

To illustrate the potential of noise-aware robust losses, we present a series of theorems that establish their superior noise tolerance compared to the original loss functions with fixed hyperparameter configurations. The detailed proofs are listed in the supplementary material Section \Rmnum{1}.

Firstly, we need to find an appropriate instance-dependent feature to assist the loss in capturing noise-awareness. Following \cite{pleiss2020identifying}, we employ the margin $m(\mathbf{x},\mathbf{y})=\tilde{f}_{\mathbf{y}}(\mathbf{x};\mathbf{w})-\max_{j\ne\mathbf{y}}\tilde{f}_{j}(\mathbf{x};\mathbf{w})$ as the instance information, where $\tilde{f}(\mathbf{x};\mathbf{w})$ represents the classifier without a softmax output layer. The margin so calculated is essentially with the property that if the margin is less than 0, the instance is considered to be classified into the wrong class, whereas a margin larger than 0 indicates a correct prediction by the classifier. In fact, our goal is to learn the underlying relationship between such instance-dependent features and noise-robust factors (hyperparameters). In addition, the margin has the property below:

\begin{proposition}
  For any $\mathbf{x}$ and its logit vector $\tilde{f}(\mathbf{x};\mathbf{w})$, assuming the largest component of logit vector is unique, there is only one $m(\mathbf{x},i)=\tilde{f}_{i}(\mathbf{x};\mathbf{w})-\max_{j\ne i}\tilde{f}_{j}(\mathbf{x};\mathbf{w})$ ($i\in\{1, \cdots, c\}$) larger than $0$, where $c$ represents the number of classes.
\end{proposition}

The conclusions presented in Thm.\ref{thm1} depend on the bounded loss conditions (as shown in Def.\ref{def3}). In order to derive the similar theories for noise-aware robust losses, we illustrate the boundedness of GCE equipped with instance-dependent hyperparameters as an example:

\begin{proposition}
  \label{pro1}
  For any $\mathbf{x}$, $q_i=q(m(\mathbf{x},i)) \in (0,1]$ and $f_i=f_{i}(\mathbf{x};\mathbf{w})$, assuming $m(\mathbf{x},j) > 0$, the sum of $\mathcal{L}_{G}$ loss with respect to all classes is bounded by:
  \begin{equation}\nonumber
    \begin{aligned}
      & C_U = \frac{c-1-(c-1)^{1-q_{min}}(1-f_j)^{q_{min}}}{q_{max}} + \frac{1-f_j^{q_j}}{q_j}\\
      &\le \sum_{i=1}^{c} \frac{1-f_{i}^{q_i}}{q_i} \le \frac{c-2+f_j}{q_{min}} + \frac{1-f_j^{q_j}}{q_j} = C_L,
    \end{aligned}    
  \end{equation}
  where $q_{min}=\min(q_1,\cdots,q_{j-1},q_{j+1},\cdots,q_{c})$, $q_{max}=\max(q_1,\cdots,q_{j-1},q_{j+1},\cdots,q_{c})$.
\end{proposition}

\textbf{Remark:} It can be observed that the smaller $C_U-C_L$, the tighter the difference between $R^{\eta}_{\mathcal{L}}(f^{*})$ and $R^{\eta}_{\mathcal{L}}(\hat{f})$, indicating that the loss is more robust (Thm.\ref{thm1}). Therefore, for GCE equipped with instance-dependent hyperparameters, we expect $$ C_U - C_L = \frac{c-2+f_j}{q_{min}} - \frac{c-1+(c-1)^{1-q_{min}}(1-f_j)^{q_{min}}}{q_{max}}$$ to be as small as possible which requires $q_{min} \rightarrow q_{max} \rightarrow 1$ \footnote[2]{We require the numerator to be as small as possible, i.e. $\max_{q_{min}} (c-1)^{1-q_{min}}(1-f_j)^{q_{min}}=(c-1)^{1-q_{min}}(1-f_j)^{q_{min}}|_{q_{min}=1}=1-f_j$, and the denominator to be as large as possible, i.e. $q_{min} \rightarrow q_{max} \rightarrow 1$.} holds for those samples with $m(\mathbf{x},i)<0$. This conclusion is consistent with the hyperparameter distribution of the noisy samples learned during the experiments (Fig.\ref{figure6} (a)). The analogous properties of other robust loss functions are shown in the supplementary material Section \Rmnum{1}. 

Finally, to facilitate a more lucid comparison between the noise-aware robust losses and the original robust losses in terms of noise tolerance, we combine the foregoing findings (Pro.\ref{pro1}) with the experimental results (Fig.\ref{figure6}), culminating in the formulation of the following theorems:

\begin{theorem}\label{thm2}
  Under the symmetric noise with $\eta \le 1 - \frac{1}{c}$, $f^*$ is the global minimizer of $R_{\mathcal{L}}(f)$ and $\hat{f}$ is the global minimizer of $R^{\eta}_{\mathcal{L}}(f)$, the following results hold: $0 \le R^{\eta}_{\mathcal{L}}(f^{*}) - R^{\eta}_{\mathcal{L}}(\hat{f}) \le \frac{\eta(C_U-C_L)}{c-1}$, and $\frac{-\eta(C_U-C_L)}{c-1-\eta c} \le R_{\mathcal{L}}(f^{*}) - R_{\mathcal{L}}(\hat{f}) \le 0$, in which:\\
  (1) for the GCE loss $\mathcal{L}_{G}$ equipped with noise-aware hyperparameters, $C_L=c-1$ and $C_U=c-2+\frac{1}{c}+\log c$;\\
  (2) for the PolySoft loss $\mathcal{L}_{P}$ equipped with noise-aware hyperparameters, $C_L=0$ and $C_U=\frac{d-1}{d}\lambda$, where $d$ and $\lambda$ are the hyperparameters of samples with $m(\mathbf{x},j)>0$;\\
  (3) the JS loss  $\mathcal{L}_{J}$ equipped with noise-aware hyperparameters, $C_L=c-1$ and $C_U=c-2+\frac{1}{c}+\log c$.\\
\end{theorem}

\begin{figure*}[t]
  \centering
  \subfigure[GCE]{\includegraphics[width=0.24\textwidth]{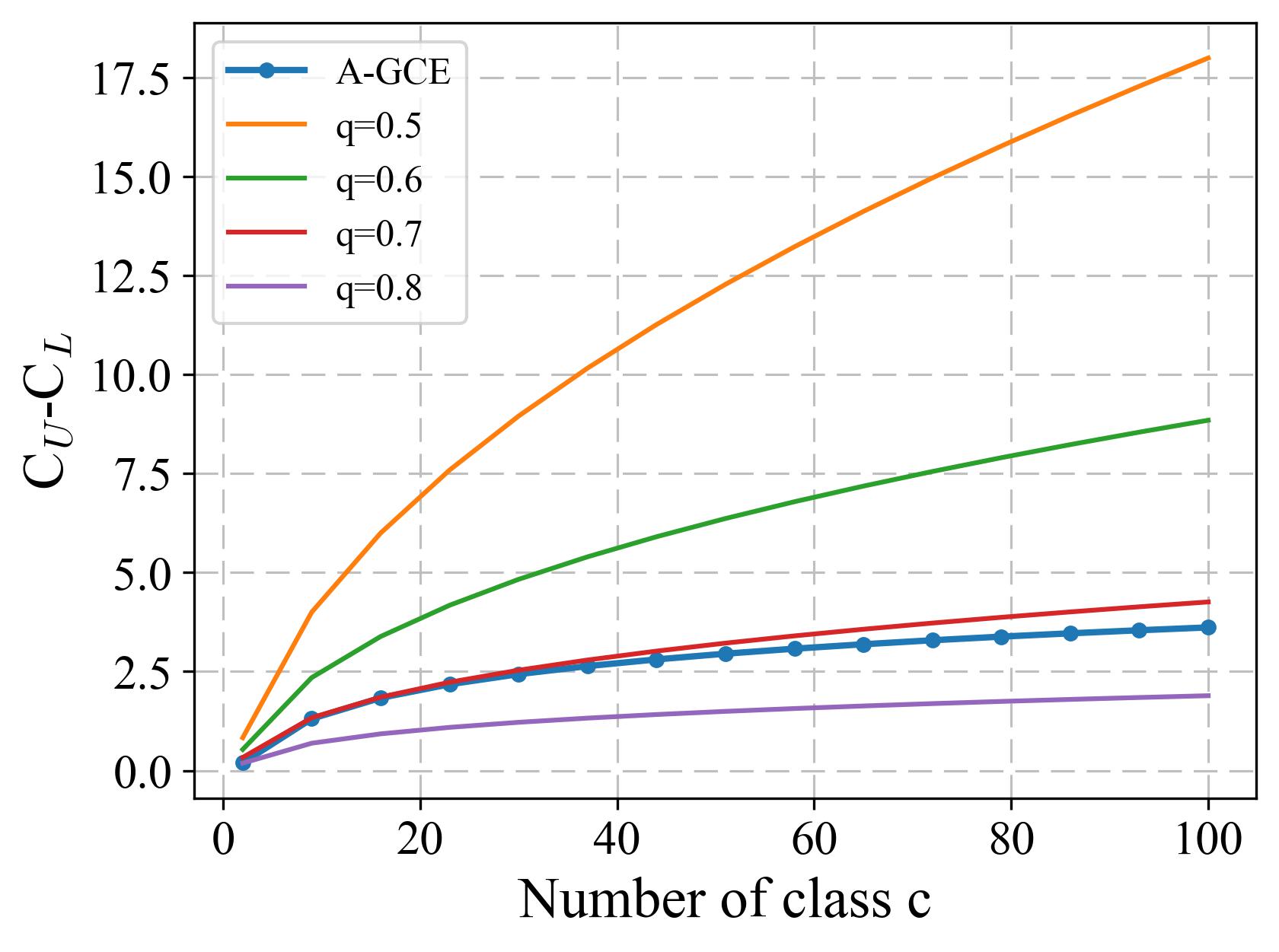}} 
  \subfigure[JS]{\includegraphics[width=0.23\textwidth]{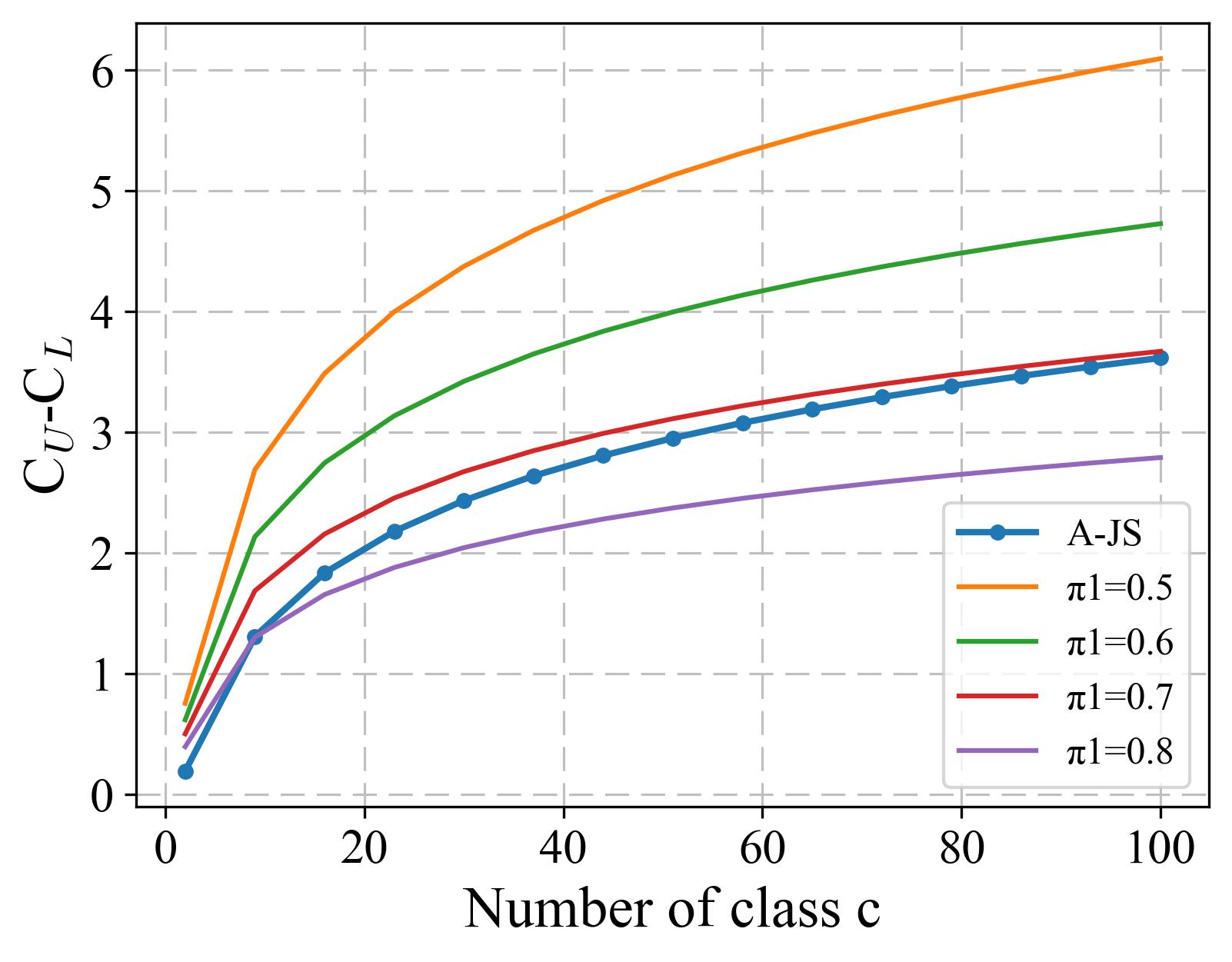}}
  \subfigure[PolySoft]{\includegraphics[width=0.235\textwidth]{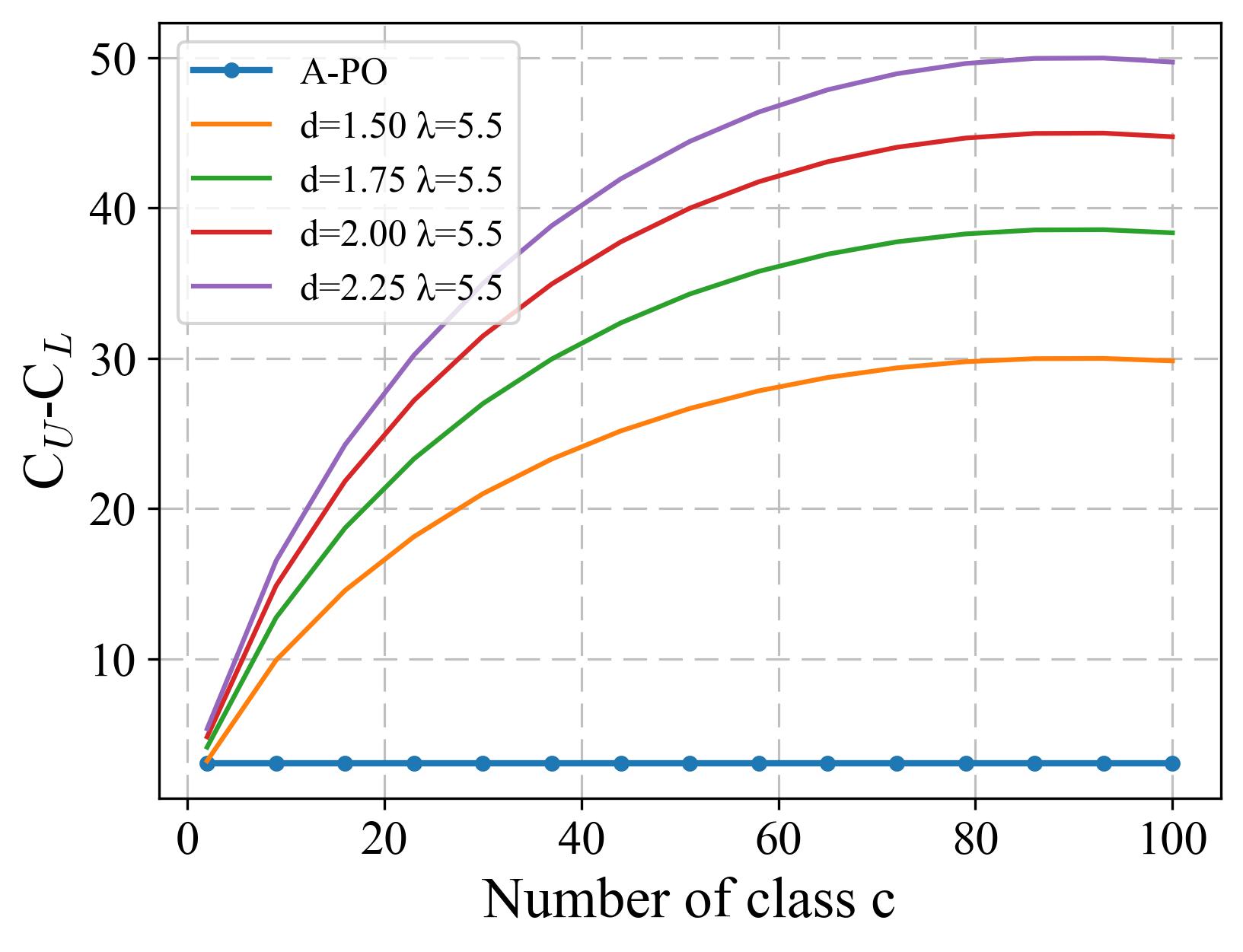}}
  \subfigure[PolySoft]{\includegraphics[width=0.235\textwidth]{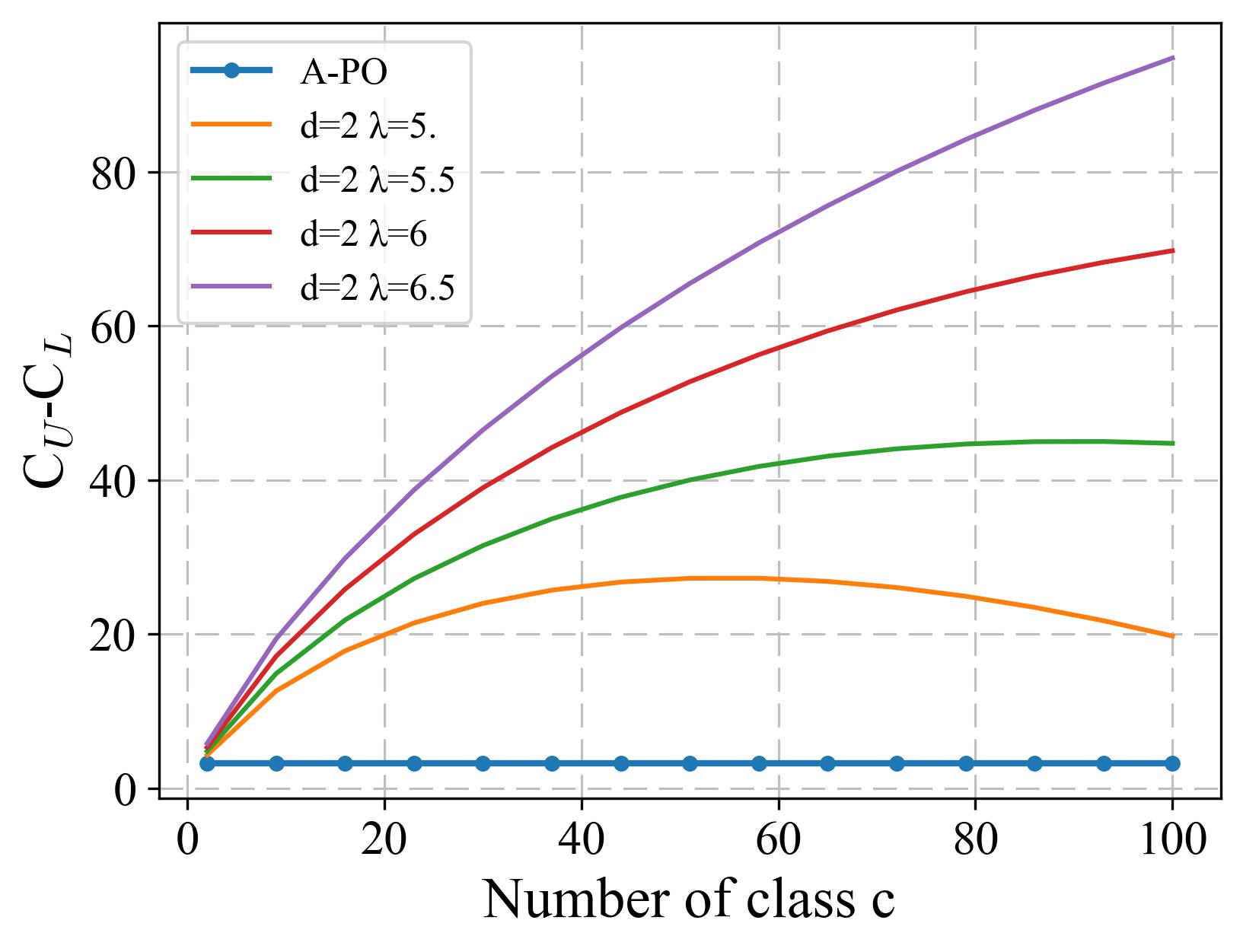}}
  \caption{The $C_U-C_L$ vs. the number of class for conventional robust losses and noise-aware robust losses. It can be inferred from Thm.\ref{thm1} that $C_U-C_L$ determines the size of $R_{\mathcal{L}}^{\eta}(f^{*})-R_{\mathcal{L}}^{\eta}(\hat{f})$'s upper bound. Here, for GCE, the comparison of $C_U-C_L$ is between $logc + \frac{1}{c}-1$ (A-GCE) and $\frac{c^{1-q}-1}{q}$ (GCE); for JS, it is between $logc + \frac{1}{c}-1$ (A-JS) and $\sum_{i=1}^{c}\mathcal{L}(\mathbf{e}_{i},\mathbf{e}_{1})-\sum_{i=1}^{c}\mathcal{L}(\mathbf{e}_{i},\mathbf{u})$ (JS), and for Polysoft, it is between $\frac{(d-1)\lambda}{d}$ (A-PolySoft)and $\frac{c(d-1)}{d}(\lambda-logc)$ (PolySoft).}
  \label{figure2}
  \vspace{0in}
\end{figure*}

Thm.\ref{thm2} demonstrate that under symmetric label noise, the $R_{\mathcal{L}}^{\eta}(f^{*})-R_{\mathcal{L}}^{\eta}(\hat{f})$ of noise-aware robust losses are bounded. In Fig.\ref{figure2}, we compare the $R_{\mathcal{L}}^{\eta}(f^{*})-R_{\mathcal{L}}^{\eta}(\hat{f})$'s upper bounds for robust losses, both with and without instance-dependent hyperparameters, and verify that noise-aware robust losses have tighter upper bounds in most multi-classification cases, indicating stronger noise tolerance against noisy labels. The theorem about asymmetric label noise is listed in supplementary material Section \Rmnum{1}.

As for SL, it cannot be proved that the difference is bounded, as SL itself is unbounded. The noise robustness properties of SL stem from the utilization of the noise-tolerant loss RCE (a component of SL). We exhibit the theories of RCE in Thm.\ref{thm1}(2) which have been proved in \cite{wang2019symmetric}.

\section{Proposed Noise-Aware-Robust-Loss-Adjuster}
\label{section4}

As previously mentioned, theoretical evidences have shown that the noise-aware robust losses exhibit promising attributes, such as enhanced noise tolerance and adaptability to diverse noise structures. Consequently, the ensuing challenge pertains to the determination of the instance-dependent hyperparameter(s) $\Lambda(\mathbf{x}_{i},\mathbf{y}_{i})$ in Eq.(\ref{eq7}) to achieve the noise-aware robust loss. In the subsequent section, we shall introduce our proposed methodology to address this matter.

\subsection{Overview of NARL-Adjuster}
{\label{section41}}

To the goal, one natural strategy is to directly learn the sample-wise hyperparameter $\Lambda(\mathbf{x}_{i},\mathbf{y}_{i})$ for robust loss. However, computational complexity of this strategy scales with the dataset size, making it infeasible especially for relatively large-scale datasets. Furthermore, based on the conclusions derived from Pro.\ref{pro1}, it can be inferred that noise-aware hyperparameters rely on the margin of each sample. The implicit relationship which is difficult to describe between sample-wise hyperparameters learned directly and noise properties hinder robust losses from improving noise-aware ability. Hence, we aim to learn the explicit mapping from individual noise information, i.e., margin, to noise-robust factors. Specifically, we apply current meta learning techniques \cite{finn2017model,hospedales2021meta,JMLR:v24:21-0742,shu2022cmw} to acquire these hyperparameter prediction functions. We call the function Noise-Aware-Robust-Loss-Adjuster (NARL-Adjuster for easy reference), and denote $h$ as NARL-Adjuster, which maps each sample's information $m(\mathbf{x},\mathbf{y})$ to its corresponding hyperparameter setting $\Lambda(\mathbf{x},\mathbf{y})$, i.e., $\Lambda(\mathbf{x},\mathbf{y}) = h(m(\mathbf{x},\mathbf{y});\Theta)$. Thus, the training objective equipped with novel robust loss can be written as:
\begin{equation}\label{eq8}
	\mathbf{w}^{*}(\Theta)= \mathop{\arg\min}_{\mathbf{w}}\frac{1}{N}\sum_{i=1}^{N}\mathcal{L}_{Robust}(f(\mathbf{x}_{i}; \mathbf{w}),\mathbf{y}_{i};h(m(\mathbf{x}_{i},\mathbf{y}_{i});\Theta)).
\end{equation}
NARL-Adjuster is hopeful to predict noise-aware hyperparameters specifically for different samples, and hence the novel robust loss is expected to be endowed with noise-aware ability of equipping different hyperparameters for clean and noisy samples, and results in better noise tolerance performance.

Now the learning variable is changed to the weight parameters of NARL-Adjuster $h$ rather than hyperparameters of robust loss functions. In other word, we need to learn the weight $\Theta$ of NARL-Adjuster $h$. Such formulation can bring the following potential merits:
\begin{itemize}
	\item The NARL-Adjuster is functioned to summarize the underlying hyperparameter setting rule. The NARL-Adjuster is potentially adaptable to different noisy labels problems (Sections \ref{section51} and \ref{section52}), since it is capable of adaptively predicting sample-wise hyperparameters conditioned on the information of samples.
	\item The learned NARL-Adjuster is plug-and-play, and can be directly deployed on unseen datasets to help predict proper hyperparameters, without extra hyperparameters in the robust loss required to be tuned (Section \ref{section55}).
	\item Once the architecture of $h$ is determined, the parameter that needs to be learned is constant and not depend on the size of dataset. This makes it easily scale to relatively larger scale training datasets (Section \ref{section55}).
\end{itemize}

\begin{figure}[t]
	\centering
	\includegraphics[width=0.42\textwidth]{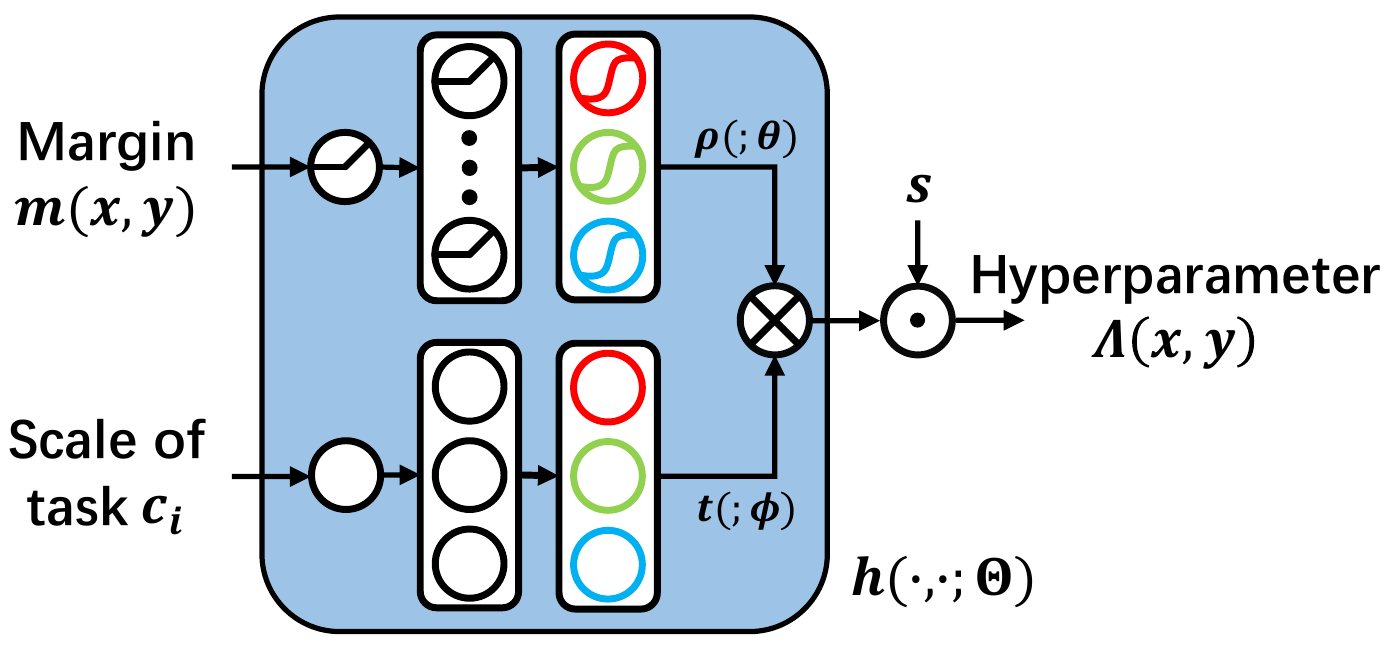}
	\caption{The structure of NARL-Adjuster.}
	\label{figure3}
\end{figure}

\subsection{Architecture and Objective Function of NARL-Adjuster}
\label{section42}

We design a meta-learner to parameterize the NARL-Adjuster, and the structure is depicted in Fig.\ref{figure3}. The NARL-Adjuster $h(m(\mathbf{x},\mathbf{y}),c_i;\Theta)$ ($\Theta=\{\theta,\phi\}$) is responsible for outputting hyperparameter(s) $\Lambda(\mathbf{x},\mathbf{y})$ for each sample. Considering that if we employ a common hyperparameter prediction function for each class, we would implicitly assume that label noise is homoscedastic, i.e. the generation of noisy labels is independent of instances and classes, such as symmetric noise. However, in real-world scenarios, label noise tends to be heterogeneous, meaning that noisy labels depend on the inputs, e.g., class/instance dependent noise. Consequently, we define the learning task as acquiring distinct hyperparameter prediction function for each class. Furthermore, considering the similarity of noise properties across different categories, we partition tasks into multiple task families and learn a hyperparameter prediction function for every task family.

The below branch of NARL-Adjuster $t(c_i;\phi)$ is used to classify samples into their respective task families based on their class information. Here, $c_i$ is the scale-related class information. Following \cite{shu2022cmw}, we denote $c_i$ as the number of instances included in each class to which the sample $\mathbf{x}$ belongs, and employ the results obtained through K-means clustering on $\{c_i\}_{i=1}^{c}$ as the below branch. In this configuration, the hidden nodes correspond to the cluster centers $\phi=\{\mu_k\}_{k=1}^{K}$ sorted in ascending order, and the output is a K-dimensional one-hot vector (i.e. task family label), whose 1-element is located at its $k$-th dimension corresponding to the nearest center $\mu_k$ to the input $c_i$.

The role of the above branch $\rho(m(\mathbf{x},\mathbf{y});\theta)$ is to fit the mapping functions of the margin and hyperparameters for each task family. We employ the margin $m(\mathbf{x},\mathbf{y})=\tilde{f}_{\mathbf{y}}(\mathbf{x};\mathbf{w})-\max_{j\ne\mathbf{y}}\tilde{f}_{j}(\mathbf{x};\mathbf{w})$ as introduced in \ref{section34}, and adopt an MLP to parameterize the above branch. At every iteration of classifier, $h(m(\mathbf{x},\mathbf{y}),c_i;\Theta)$ can learn an explicit instance-hyperparameter dependent relationship, such that the NARL-Adjuster can predict noise-aware hyperparameter(s) relying on the evolving margin of samples. The whole forward computing process can be written as:
$$\Lambda(\mathbf{x},\mathbf{y})=s \odot \rho(m(\mathbf{x},\mathbf{y});\theta)^{T} \otimes t(c_i;\phi)$$
where $\odot$ denotes the element-wise product and $\otimes$ denotes the matmul product. $s$ is column vector whose dimension is the same as the number of hyperparameters included in the robust loss function to control the range of hyperparameters. As for the structure of $\rho(m(\mathbf{x},\mathbf{y});\theta)$, we set the hidden layer containing 100 nodes with ReLU activation function. For each task family, we use a fully connected layer whose inputs are the outputs of hidden layer and sigmoid outputs are hyperparameters of robust loss. We expect to utilize the approximation ability of the meta-learner to learn the noise-aware hyperparameter prediction function.

The parameters contained in NARL-Adjuster can be learned by using the general meta-learning solving techniques \cite{schmidhuber1992learning,thrun2012learning,finn2017model,shu2019meta,shu2018small,hospedales2021meta}. Specifically, assume that we have a small amount meta-data set (i.e., with clean labels) $D_{meta}=\{\mathbf{x}_{i}^{(m)},\mathbf{y}_{i}^{(m)}\}_{i=1}^{M}$, representing the meta-knowledge of groundtruth sample-label distribution, where $M$ is the number of meta-samples, and $M\ll N$. We can then formulate a meta-loss minimization problem with respect to $\Theta=\{\theta,\phi\}$ as:
\begin{equation}\label{eq9}
	\begin{aligned}
		\Theta^{*}&=\arg\min_{\Theta} \frac{1}{M}\sum_{i=1}^{M}\mathcal{L}_{Meta}(f(\mathbf{x}_i^{(m)};\mathbf{w}^{*}(\Theta)),\mathbf{y}_{i}^{(m)}),
	\end{aligned}
\end{equation}
where $\mathcal{L}_{Meta}$ usually uses the CE loss calculated on meta-data.

\begin{algorithm}[t]
	%\textsl{}\setstretch{1.8}
	\renewcommand{\algorithmicrequire}{\textbf{Input:}}
	\renewcommand{\algorithmicensure}{\textbf{Output:}}
	\caption{The Meta-Train Algorithm of NARL-Adjuster}
	\label{alg1}
	\begin{algorithmic}[1]
    \REQUIRE Training data $\hat{D}$, meta-data set $D_{Meta}$, batch size $n, m$, max iterations $T$, updating period $T_{meta}$.
    \ENSURE Classifier network parameter $\mathbf{w}^{(T)}$ and NARL-Adjuster parameter $\theta^{(T)}$, $k\in \mathcal{K}\subset\{1,...,T\}$.
    \STATE Apply K-means on the number of samples of all classes to obtain $\phi^{*}$.
		\STATE Initialize classifier network parameter $\mathbf{w}^{(0)}$ and NARL-Adjuster parameter $\theta^{(0)}$.
        \FOR{$t=0$ to $T-1$}
            \STATE $\{\mathbf{x},\mathbf{y}\}\gets$ SampleMiniBatch$(\hat{D},n)$.
            \IF{$t$ \% $T_{Meta}=0$,}
               \STATE Calculate $\tilde{\mathbf{w}}^{(t)}(\theta)$ by Eq.(\ref{eq11}).
               \STATE$\{\mathbf{x}^{(m)},\mathbf{y}^{(m)}\}\gets$ SampleMiniBatch$(D_{Meta},m)$.
               \STATE Update $\theta^{(t+1)}$ by Eq.(\ref{eq10}).
            \ENDIF
            \STATE Update $\mathbf{w}^{(t+1)}$ by Eq.(\ref{eq12}).
        \ENDFOR
	\end{algorithmic}  
\end{algorithm}

\subsection{Learning Algorithm of NARL-Adjuster}
\label{section43}

\subsubsection{Meta-Train: Summarizing the underlying hyperparameter setting rule}

The NARL-Adjuster can be meta-trained to improve the noise tolerance ability of robust loss by solving the bilevel optimization problem as Eq.(\ref{eq8}) and Eq.(\ref{eq9}). The objective function is the standard bi-level optimization problem \cite{franceschi2018bilevel}, where Eq.(\ref{eq8}) and (\ref{eq9}) are inner and outer optimization problems, respectively. It is expensive to obtain the exact solution of $\mathbf{w}^{*}$ and $\Theta^{*}$ by calculating two nested loops of optimization. We decide to first solve the K-means to obtain the parameter $\phi^{*}$, and then, we adopt an online approximation strategy \cite{finn2017model,shu2019meta} to jointly update parameters of both the classifier network and $\rho(m(\mathbf{x},\mathbf{y});\theta)$ in an iterative manner to guarantee the efficiency of the algorithm. Concretely, the updating process is as follows:

\begin{figure*}[t]
	\centering
	\includegraphics[width=1.\textwidth]{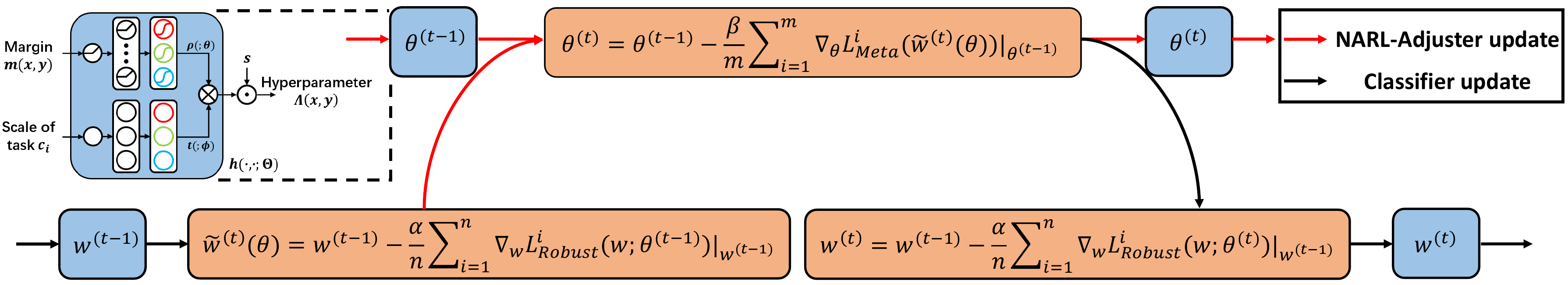}
	\caption{Main flowchart of the proposed NARL-Adjuster meta-train algorithm.}
	\label{figure4}
\end{figure*}

\textbf{Updating $\theta$}. At iteration step $t$, we need to update NARL-Adjuster parameter $\theta^{(t)}$ on the basis of the classifier parameter $\mathbf{w}^{(t-1)}$ and $\theta^{(t-1)}$ obtained in the last iteration by minimizing the meta loss defined in Eq.(\ref{eq8}). SGD is employed to optimize the $\theta$ on $m$ mini-batch samples ${D}_{m}$ from $D_{meta}$, i.e.,
\begin{equation}\label{eq10}
  \theta^{(t)}=\theta^{(t-1)}-\frac{\beta}{m}\sum_{i=1}^{m}\nabla_{\theta}L_{Meta}^{i}(\tilde{\mathbf{w}}^{(t)}(\theta))|_{\theta^{(t-1)}},
\end{equation}
where the following equation is used to formulate $\tilde{\mathbf{w}}^{(t)}$ on $n$ mini-batch training samples $D_{n}$ from $\hat{D}$:
\begin{equation}\label{eq11}
  \tilde{\mathbf{w}}^{(t)}(\theta) = \mathbf{w}^{(t-1)} -\frac{\alpha}{n}\sum_{i=1}^{n} \nabla_{\mathbf{w}}L_{Robust}^{i}(\mathbf{w};\theta^{(t-1)})|_{\mathbf{w}^{(t-1)}},
\end{equation}
where $\alpha$, $\beta$ are the step sizes, $L_{Meta}^{i}(\tilde{\mathbf{w}}^{(t)}(\theta)) =\mathcal{L}_{Meta}(f(\mathbf{x}_i^{(m)};\tilde{\mathbf{w}}^{(t)}(\theta)),\mathbf{y}_{i}^{(m)}) $ and $ L_{Robust}^{i}(\mathbf{w};\theta) = \mathcal{L}_{Robust}(f(\mathbf{x}_{i}; \mathbf{w}),\mathbf{y}_{i};h(m(\mathbf{x}_{i},\mathbf{y}_{i}),c_i;\theta,\phi^{*}))$.

\textbf{Updating $\mathbf{w}$}.
With the NARL-Adjuster parameter $\theta^{(t)}$, the classifier parameters $\mathbf{w}^{(t)}$ can then be updated by:
\begin{equation}\label{eq12}
  \mathbf{w}^{(t)}=\mathbf{w}^{(t-1)}-\frac{\alpha}{n} \sum_{i=1}^{n}\nabla_{\mathbf{w}}L_{Robust}^{i}(\mathbf{w};\theta^{(t)})|_{\mathbf{w}^{(t-1)}}.
\end{equation}

The algorithm in the meta-train stage can then be summarized in Algorithm \ref{alg1}, and the diagram is shown in Fig.\ref{figure4}. Considering the step-wise optimization for $\theta$ is expensive, we attempt to update $\theta$ once after updating $\mathbf{w}$ several steps ($T_{Meta}$). All computations of gradients can be efficiently implemented by automatic differentiation techniques and easily generalized to any deep learning architectures. The algorithm can be easily implemented using popular deep learning frameworks like PyTorch \cite{paszke2017automatic}. The algorithm can then be easily integrated with any robust loss functions to improve their noise tolerance property. Specifically, we denote algorithms integrated with robust losses defined in Eq.(\ref{eq2}), (\ref{eq3}), (\ref{eq4}) and (\ref{eq5}) as A-GCE, A-SL, A-PolySoft and A-JS algorithms, respectively.

\begin{algorithm}[!t]
	%\textsl{}\setstretch{1.8}
	\renewcommand{\algorithmicrequire}{\textbf{Input:}}
	\renewcommand{\algorithmicensure}{\textbf{Output:}}
	\caption{The Meta-Test Algorithm of NARL-Adjuster}
	\label{alg2}
	\begin{algorithmic}[1]
	    \REQUIRE Training data $D_{New}$, batch size $n$, max iterations $T$, meta-learned NARL-Adjuster $h(\cdot, \cdot;\theta^{(k)}), k \in K$.
	    \ENSURE Classifier network parameter $\xi^{(T)}$.
      \STATE Apply K-means on the number of samples of all classes to obtain $\phi^{*}_{new}$.
        \FOR{$t=0$ to $T-1$}
            \STATE $\{\mathbf{x},\mathbf{y}\}\gets$ SampleMiniBatch$(D_{New},n)$.
            \STATE Compute the margin $m(\mathbf{x},\mathbf{y})$, and then NARL-Adjuster predicts the instance-dependent hyperparameters $h(m(\mathbf{x}_{i},\mathbf{y}_{i}),c^{new}_i;\theta,\phi^{*}_{new})$ of robust loss for current iteration.
            \STATE Update $\xi^{(t+1)}$ by Eq.(\ref{eq12}).
        \ENDFOR
	\end{algorithmic}  
\end{algorithm}

\subsubsection{Meta-Test: Deploying the learned NARL-Adjuster on unseen datasets}

The meta-learned NARL-Adjuster has extracted the hyperparameter setting rule, which is expected to set hyperparameters of robust loss on novel noisy label classification tasks without extra robust loss hyperparameters required to be tuned. In order to better maintain the hyperparameter adjusting dynamics in the meta-train stage, we more prefer to restore several NARL-Adjuster forms with parameters $\theta^{(k)}$, $k\in \mathcal{K}\subset\{1,...,T\}$ (e.g., $\theta^{(\frac{T}{3})}, \theta^{(\frac{2T}{3})},\theta^{(T)}$ are employed in our experiments) and dynamically adopt specific ones along different range of classifier training stage in Meta-Test. Hence the parameters $\xi$ of new classifier network are updated by,
\begin{equation}\label{eq13}
	\xi^{(t+1)} = \xi^{(t)} - \frac{\eta}{n}\sum_{i=1}^{n}\nabla_{\xi} L_{Robust}^{i}(\xi;\theta^{(k)}) |_{\xi^{(t)}},
\end{equation}
where $\theta^{(k)}$ ($k\in \mathcal{K}$) represents the parameters of the subset of the meta-trained NARL-Adjusters, $L_{Robust}^{i}(\xi;\theta^{(k)})=\mathcal{L}_{Robust}(f(\mathbf{x}_{i}; \xi),\mathbf{y}_{i};h(m(\mathbf{x}_{i},\mathbf{y}_{i}),c^{new}_i;\theta,\phi^{*}_{new}))$ and $\phi^{*}_{new}$ is obtained through the implementation of the K-means algorithm on the number of samples of each class in the new classification task. The whole meta-test process refers to Algorithm \ref{alg2}. It can be seen that the meta-learned NARL-Adjuster is hopeful to be useful in practical problems as it requires negligible increase in the parameter size and small transferrable cost for new tasks compared with conventional robust loss methods with hyperparameters required to be manually pre-specified.

\section{Experiments}
\label{section5}

To evaluate the capability of the NARL-Adjuster algorithm, firstly, we conduct experiments on CIFAR-10 and CIFAR-100 datasets with synthetic symmetric, asymmetric (Section \ref{section51}), and instance-dependent label noises (Section \ref{section52}) to justify whether our approach can learn suitable hyperparameter prediction functions under various noise structures. Secondly, we compare the proposed algorithm with SOTA methods on CIFAR-10, CIFAR-100, and mini-WebVision datasets (Section \ref{section53}). Then, we analyze how NARL-Adjuster improves noise tolerance of robust loss functions in Section \ref{section54}. Finally, we transfer the meta-learned NARL-Adjuster to unseen datasets with noisy labels, such as relatively large-scale datasets TinyImageNet and real-world noisy datasets ANIMAL-10N and Food-101N, demonstrating its transferability in Section \ref{section55}. The experimental setups can be found in supplementary material \Rmnum{2}.

\begin{table*}[!t]
  \renewcommand\arraystretch{1.2}
  \centering
  \caption{Test accuracy (\%) of competing methods on CIFAR-10, CIFAR-100 under instance-independent noise. The best results are in bold.}
  \label{table1}
  \resizebox{\linewidth}{!}{
  \begin{tabular}{c|c|l|l|l|l|l|l}
  \hline
  
  \multirow{2}{*}{Datasets} & \multirow{2}{*}{Methods} & \multicolumn{4}{c|}{Symmetric Noise} & \multicolumn{2}{c}{Asymmetric Noise }\\
  \cline{3-8}
  
  & & \multicolumn{1}{c|}{0} & \multicolumn{1}{c|}{0.2} & \multicolumn{1}{c|}{0.4} & \multicolumn{1}{c|}{0.6} & \multicolumn{1}{c|}{0.2} & \multicolumn{1}{c}{0.4}\\
  \hline
  \hline
  
  \multirow{11}{*}{CIFAR-10} & CE & \textbf{93.11} \bm{$\pm$} \textbf{0.11} & 87.77 $\pm$ 0.10 & 84.13 $\pm$ 0.27 & 78.90 $\pm$ 0.34 & 88.93 $\pm$ 0.24 & 82.18 $\pm$ 0.92 \\ 
  
  & Forward & 92.75 $\pm$ 0.08 & 88.72 $\pm$ 0.09 & 84.94 $\pm$ 0.27 & 79.33 $\pm$ 0.29 & 89.30 $\pm$ 0.22 & 85.68 $\pm$ 0.55 \\ 
  
  & MW-Net & 92.85 $\pm$ 0.17 & 90.09 $\pm$ 0.38 & 87.27 $\pm$ 0.34 & 82.50 $\pm$ 0.53 & 90.80 $\pm$ 0.32 & 87.27 $\pm$ 0.58 \\ 
  \cline{2-8}
  
  & GCE & 91.11 $\pm$ 0.13 & 89.87 $\pm$ 0.05 & 87.19 $\pm$ 0.29 & 83.16 $\pm$ 0.19 & 89.82 $\pm$ 0.13 & 83.12 $\pm$ 0.94 \\ 
  
  & {\color{blue}A-GCE} & 92.88 $\pm$ 0.16\scriptsize{\color{red}{$\uparrow$1.77}} & 90.72 $\pm$ 0.12\scriptsize{\color{red}{$\uparrow$0.85}} & \textbf{88.19} \bm{$\pm$} \textbf{0.13}\scriptsize{\color{red}{$\uparrow$1.00}} & 83.36 $\pm$ 0.14\scriptsize{\color{red}{$\uparrow$0.20}} & \textbf{91.56} \bm{$\pm$} \textbf{0.12}\scriptsize{\color{red}{$\uparrow$1.74}} & \textbf{88.08} \bm{$\pm$} \textbf{0.28}\scriptsize{\color{red}{$\uparrow$4.96}} \\ 
  \cline{2-8}
  
  & SL & 89.37 $\pm$ 0.12 & 88.51 $\pm$ 0.19 & 86.38 $\pm$ 0.24 & 82.07 $\pm$ 0.36 & 88.20 $\pm$ 0.13 & 81.93 $\pm$ 0.49 \\ 
  
  & {\color{blue}A-SL} & 91.51 $\pm$ 0.20\scriptsize{\color{red}{$\uparrow$2.14}} & 89.96 $\pm$ 0.10\scriptsize{\color{red}{$\uparrow$1.45}} & 87.07 $\pm$ 0.28\scriptsize{\color{red}{$\uparrow$0.69}} & 82.66 $\pm$ 0.20\scriptsize{\color{red}{$\uparrow$0.59}} & 90.34 $\pm$ 0.26\scriptsize{\color{red}{$\uparrow$2.14}} & 83.41 $\pm$ 0.77\scriptsize{\color{red}{$\uparrow$1.48}} \\ 
  \cline{2-8}
  
  & PolySoft & 91.84 $\pm$ 0.05 & 90.11 $\pm$ 0.11 & 86.74 $\pm$ 0.08 & 79.25 $\pm$ 0.16 & 89.47 $\pm$ 0.21 & 86.18 $\pm$ 0.24 \\ 
  
  & {\color{blue}A-PolySoft} & 92.95 $\pm$ 0.13\scriptsize{\color{red}{$\uparrow$1.11}} & 90.71 $\pm$ 0.16\scriptsize{\color{red}{$\uparrow$0.60}} & 87.80 $\pm$ 1.06\scriptsize{\color{red}{$\uparrow$0.99}} & \textbf{83.65} \bm{$\pm$} \textbf{1.06}\scriptsize{\color{red}{$\uparrow$4.40}} & 91.16  $\pm$ 0.18\scriptsize{\color{red}{$\uparrow$1.69}} & 87.55 $\pm$ 0.53\scriptsize{\color{red}{$\uparrow$1.37}} \\ 
  \cline{2-8}
  
  & JS & 92.72 $\pm$ 0.05 & 89.75 $\pm$ 0.15 & 87.62 $\pm$ 0.08 & 82.78 $\pm$ 0.17 & 90.27 $\pm$ 0.16 & 86.65 $\pm$ 0.30 \\ 
  
  & {\color{blue}A-JS} & 92.84 $\pm$ 0.15\scriptsize{\color{red}{$\uparrow$0.12}} & \textbf{90.73} \bm{$\pm$} \textbf{0.03}\scriptsize{\color{red}{$\uparrow$0.98}} & 88.11 $\pm$ 0.12\scriptsize{\color{red}{$\uparrow$0.49}} & 83.02 $\pm$ 0.18\scriptsize{\color{red}{$\uparrow$0.24}} & 90.99 $\pm$ 0.21\scriptsize{\color{red}{$\uparrow$0.72}} & 87.98 $\pm$ 0.34\scriptsize{\color{red}{$\uparrow$1.33}} \\ 
  
  \hline
  
  \multirow{11}{*}{CIFAR-100} & CE & 70.56 $\pm$ 0.17 & 62.62 $\pm$ 0.36 & 55.22 $\pm$ 0.46 & 44.00 $\pm$ 0.47 & 64.25 $\pm$ 0.32 & 59.46 $\pm$ 0.28 \\ 
  
  & Forward & 69.31 $\pm$ 0.21 & 62.56 $\pm$ 0.15 & 55.71 $\pm$ 0.12 & 44.70 $\pm$ 0.57 & 64.50 $\pm$ 0.21 & 60.02 $\pm$ 0.16 \\ 
  
  & MW-Net & 70.74 $\pm$ 0.17 & 66.58 $\pm$ 0.34 & 59.07 $\pm$ 0.57 & 55.47 $\pm$ 0.45 & 65.72 $\pm$ 0.16 & 60.71 $\pm$ 0.10 \\ 
  \cline{2-8}
  
  & GCE & 68.49 $\pm$ 0.17 & 65.78 $\pm$ 0.22 & 62.39 $\pm$ 0.38 & 52.86 $\pm$ 0.41 & 65.71 $\pm$ 0.18 & 60.11 $\pm$ 0.27 \\ 
  
  & {\color{blue}A-GCE} & 70.99 $\pm$ 0.26\scriptsize{\color{red}{$\uparrow$2.50}} & 67.22 $\pm$ 0.31\scriptsize{\color{red}{$\uparrow$1.44}} & 62.83 $\pm$ 0.13\scriptsize{\color{red}{$\uparrow$0.44}} & 55.95 $\pm$ 0.15\scriptsize{\color{red}{$\uparrow$3.09}} & 66.39 $\pm$ 0.08\scriptsize{\color{red}{$\uparrow$0.68}} & 62.14 $\pm$ 0.28\scriptsize{\color{red}{$\uparrow$2.03}} \\ 
  \cline{2-8}
  
  & SL & 67.80 $\pm$ 0.49 & 59.86 $\pm$ 0.45 & 53.96 $\pm$ 0.66 & 44.61 $\pm$ 0.84 & 60.86 $\pm$ 0.61 & 57.74 $\pm$ 0.23 \\ 
  
  & {\color{blue}A-SL} & 70.38 $\pm$ 0.12\scriptsize{\color{red}{$\uparrow$2.58}} & 65.26 $\pm$ 0.22\scriptsize{\color{red}{$\uparrow$5.40}} & 58.71 $\pm$ 0.62\scriptsize{\color{red}{$\uparrow$4.75}} & 48.59 $\pm$ 0.31\scriptsize{\color{red}{$\uparrow$3.98}} & 65.52 $\pm$ 0.24\scriptsize{\color{red}{$\uparrow$4.66}} & 57.75 $\pm$ 0.32\scriptsize{\color{red}{$\uparrow$0.01}} \\ 
  \cline{2-8}
  
  & Ployoft & 69.60 $\pm$ 0.07 & 65.73 $\pm$ 0.20 & 62.90 $\pm$ 0.10 & 54.64 $\pm$ 0.26 & 63.89 $\pm$ 0.22 & 61.08 $\pm$ 0.13 \\ 
  
  & {\color{blue}A-PloySoft} & 70.83 $\pm$ 0.12\scriptsize{\color{red}{$\uparrow$1.23}} & 67.70 $\pm$ 0.18\scriptsize{\color{red}{$\uparrow$1.97}} & \textbf{63.64} $\pm$ \textbf{0.16}\scriptsize{\color{red}{$\uparrow$0.74}} & \textbf{56.82} $\pm$ \textbf{0.25}\scriptsize{\color{red}{$\uparrow$2.18}} & \textbf{66.61} $\pm$ \textbf{0.13}\scriptsize{\color{red}{$\uparrow$2.72}} & 62.30 $\pm$ 0.22\scriptsize{\color{red}{$\uparrow$1.22}} \\ 
  \cline{2-8}
  
  & JS & 70.56 $\pm$ 0.10 & 67.72 $\pm$ 0.23 & 62.74 $\pm$ 0.18 & 54.48 $\pm$ 0.33 & 61.81 $\pm$ 0.08 & 58.63 $\pm$ 0.18 \\ 
  
  & {\color{blue}A-JS} & \textbf{71.18} \bm{$\pm$} \textbf{0.14}\scriptsize{\color{red}{$\uparrow$0.62}} & \textbf{68.25} \bm{$\pm$} \textbf{0.09}\scriptsize{\color{red}{$\uparrow$0.53}} & 63.48 \bm{$\pm$} 0.18\scriptsize{\color{red}{$\uparrow$0.74}} & 56.22 \bm{$\pm$} 0.24\scriptsize{\color{red}{$\uparrow$1.74}} & 66.54 \bm{$\pm$} 0.14\scriptsize{\color{red}{$\uparrow$4.73}} & \textbf{62.92} \bm{$\pm$} \textbf{0.26}\scriptsize{\color{red}{$\uparrow$4.29}} \\ 
  \hline
  
  \end{tabular}}
\end{table*}

\subsection{Instance-Independent Label Noise}
\label{section51}

\textbf{Datasets.} We first verify the effectiveness of our NARL-Adjuster to learn proper hyperparameter prediction functions on two benchmark datasets: CIFAR-10 and CIFAR-100 \cite{krizhevsky2009learning}, consisting of 32 × 32 color images arranged in 10 and 100 classes, respectively. Both datasets contain 50,000 training and 10,000 test images. We randomly select 1,000 clean images in the validation set as meta data. These datasets are popularly used for evaluation of learning with noisy labels in the previous literatures \cite{reed2014training,patrini2017making,goldberger2016training}. We test two types of label noise: symmetric noise and asymmetric (class dependent) noise. Symmetric noisy labels are generated by flipping the labels of a given proportion of training samples to one of the other class labels uniformly \cite{zhang2021understanding}. For asymmetric noisy labels, we use the setting in \cite{patrini2017making}, generating the label noise by mapping TRUCK $\to$ AUTOMOBILE, BIRD $\to$ AIRPLANE, DEER $\to$ HORSE, and CAT $\to$ DOG with a given probability for CIFAR-10. Also, we consider a more realistic hierarchical corruption in CIFAR-100 as described in \cite{hendrycks2018using}, which applies uniform corruption only to semantically similar classes.

\textbf{Baselines.} We compare NARL-Adjuster algorithm with the following typical methods on handling noisy label datasets, and implement all methods with default settings in the original paper by PyTorch. 1) \textbf{CE}, which uses CE loss to train the DNNs on datasets with noisy labels. 2) \textbf{Forward} \cite{patrini2017making}, which corrects the prediction by the label transition matrix. 3) \textbf{Meta-Weight-Net} \cite{shu2019meta}, which uses an MLP net to learn the weighting function in a data-driven fashion, representing the SOTA sample weighting methods. 4) \textbf{PolySoft} \cite{gong2018decomposition}, 5) \textbf{GCE} \cite{zhang2018generalized}, 6) \textbf{SL} \cite{wang2019symmetric}, 7) \textbf{JS} \cite{englesson2021generalized} are four representative robust loss methods. The meta data in these methods are used as validation set for cross-validation to search the best hyperparameters except for Meta-Weight-Net.

\begin{figure}[t]
	\centering
  \setlength{\abovecaptionskip}{-0.25cm}
	\includegraphics[width=0.36\textwidth]{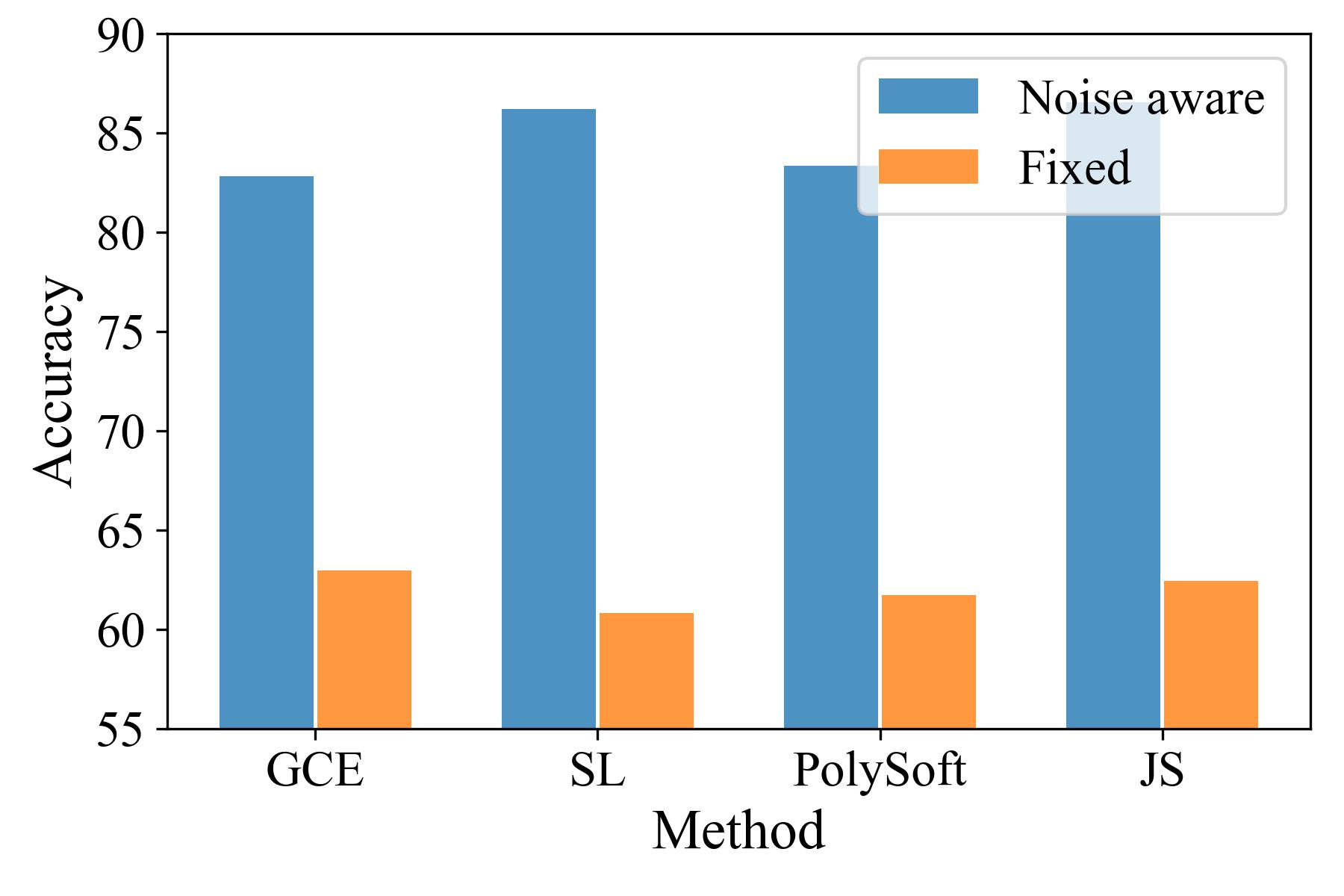}
	\caption{Performance comparison of robust losses and robust losses with noise-aware hyperparameters on CIFAR-10 dataset under 60\% asymmetric noise.}
	\label{figure5}
  \vspace{-0.63cm}
\end{figure}

\textbf{Results.} The classification accuracy of CIFAR-10 and CIFAR-100 under symmetric and asymmetric noise cases are reported in Table \ref{table1} and we use the average results of last 5 epochs. As can be seen, our NARL-Adjuster algorithm (in color blue) gets a performance gain on the original algorithm for almost all noise rates, noise types on all datasets (as indicated by the footnotes). From Table \ref{table1}, it can be observed that: 1) Compared with other methods, A-JS has the best performances or the second-best performance on CIFAR-100 under various noisy setting. 2) SL drops more quickly on CIFAR-100 than CIFAR-10, and A-SL improves the accuracy more evidently, around 5\% points on CIFAR-100 under several noise rates. 3) A-PolySoft outperforms the SOTA sample reweighting method Meta-Weight-Net, possibly caused by its monotonically decreasing form of robust loss, making it noise robust, as illustrated in supplementary material Section \Rmnum{3}. 4) The proposed approach demonstrates enhanced efficacy in handling heterogenous noise. As depicted in Fig.\ref{figure5}, on the CIFAR-10 dataset with 60\% asymmetric noise, our method exhibits an improvement exceeding 20\% compared to the robust loss methods. This advancement can be attributed to the structure of NARL-Adjuster (Fig.\ref{figure3}), which is able to learn different hyperparameter prediction functions for classes exhibiting diverse noise properties. 5) When dealing with different types of label noise, the performance of original robust losses fluctuate significantly (e.g., the noise type change from symmetric to asymmetric), but robust losses with instance-dependent hyperparameters can maintain relatively low performance degradation. 6) With the noise rate increasing, the performance of conventional robust losses drops quickly and the proposed method can mitigate this phenomenon to some extent, especially on CIFAR-100. Both 5) and 6) empirically validate the proposed theoretical perspectives (Thm.\ref{thm2}): the noise-aware robust losses exhibit enhanced noise tolerance on different noise structure forms.

\begin{table*}[t]
  \renewcommand\arraystretch{1.2}
  \centering
  \caption{Test accuracy (\%) of competing methods on CIFAR-10, CIFAR100 under instance-dependent noise. The best results are in bold.}
  \label{table2}
  %\resizebox{\tabcolsep}{}
  {
  \begin{tabular}{c|c|l|l|l|l}
  \hline
  
  \multirow{2}{*}{Dataset} & \multirow{2}{*}{Methods} & \multicolumn{4}{c}{Instances Noise}\\
  \cline{3-6}
  
  & & \multicolumn{1}{c|}{0.2} & \multicolumn{1}{c|}{0.3} & \multicolumn{1}{c|}{0.4} & \multicolumn{1}{c}{0.5} \\
  \hline
  \hline
  
  \multirow{11}{*}{CIFAR-10} & CE & 75.37 $\pm$ 0.65 & 65.08 $\pm$ 0.46 & 55.02 $\pm$ 0.57 & 43.62 $\pm$ 0.81\\ 
  
  & Forward & 75.68 $\pm$ 0.48 & 66.17 $\pm$ 0.50 & 54.83 $\pm$ 0.93 & 43.86 $\pm$ 0.32\\
  
  & MW-Net & 85.35 $\pm$ 0.15 & 78.64 $\pm$ 0.25 & \textbf{71.84} \bm{$\pm$} \textbf{0.53} & 44.81 $\pm$ 2.85\\
  \cline{2-6}
  
  & GCE & 84.31 $\pm$ 0.21 & 74.15 $\pm$ 0.53 & 57.41 $\pm$ 0.34 & 43.18 $\pm$ 1.24\\ 
  
  & {\color{blue}A-GCE} & 85.18 $\pm$ 0.17\scriptsize{\color{red}{$\uparrow$0.87}} & 78.67 $\pm$ 0.18\scriptsize{\color{red}{$\uparrow$4.52}} & 64.22 $\pm$ 0.25\scriptsize{\color{red}{$\uparrow$6.81}} & 45.74 $\pm$ 0.92\scriptsize{\color{red}{$\uparrow$2.56}}\\ 
  \cline{2-6}
  
  & SL & 83.16 $\pm$ 0.43 & 74.01 $\pm$ 0.60 & 58.34 $\pm$ 0.52 & 41.79 $\pm$ 0.62\\ 
  
  & {\color{blue}A-SL} & \textbf{86.50} \bm{$\pm$} \textbf{0.15}\scriptsize{\color{red}{$\uparrow$2.34}} & \textbf{82.20} \bm{$\pm$} \textbf{0.31}\scriptsize{\color{red}{$\uparrow$8.19}} & 62.82 $\pm$ 0.32\scriptsize{\color{red}{$\uparrow$4.48}} & 42.91 $\pm$ 0.38\scriptsize{\color{red}{$\uparrow$1.12}}\\ 
  \cline{2-6}
  
  & PolySoft & 77.25 $\pm$ 0.39 & 66.95 $\pm$ 0.52 & 54.87 $\pm$ 0.92 & 44.01 $\pm$ 1.00\\ 
  
  & {\color{blue}A-PolySoft} & 84.28 $\pm$ 0.19\scriptsize{\color{red}{$\uparrow$7.03}} & 78.64 $\pm$ 0.18\scriptsize{\color{red}{$\uparrow$11.69}} & 70.47 $\pm$ 0.31\scriptsize{\color{red}{$\uparrow$15.60}} & \textbf{54.89} \bm{$\pm$} \textbf{0.54}\scriptsize{\color{red}{$\uparrow$10.88}}\\ 
  \cline{2-6}
  
  & JS & 83.31 $\pm$ 0.18 & 72.09 $\pm$ 0.35 & 60.59 $\pm$ 1.24 & 43.05 $\pm$ 0.40\\ 
  
  & {\color{blue}A-JS} & 84.84 $\pm$ 0.23\scriptsize{\color{red}{$\uparrow$1.53}} & 78.40 $\pm$ 0.17\scriptsize{\color{red}{$\uparrow$6.31}} & 63.82 $\pm$ 0.41\scriptsize{\color{red}{$\uparrow$3.23}} & 44.31 $\pm$ 0.41\scriptsize{\color{red}{$\uparrow$1.26}}\\ 
  \hline
  
  \multirow{11}{*}{CIFAR-100} & CE & 57.08 $\pm$ 0.11 & 48.60 $\pm$ 0.10 & 38.95 $\pm$ 0.24 & 29.38 $\pm$ 0.22\\ 
  
  & Forward & 57.88 $\pm$ 0.16 & 48.76 $\pm$ 0.27 & 39.32 $\pm$ 0.27 & 29.81 $\pm$ 0.50\\
  
  & MW-Net & 60.74 $\pm$ 0.17 & 52.48 $\pm$ 0.18 & 40.10 $\pm$ 0.42 & 30.25 $\pm$ 0.31\\
  \cline{2-6}
  
  & GCE & 58.77 $\pm$ 0.14 & 48.85 $\pm$ 0.20 & 38.74 $\pm$ 0.13 & 28.59 $\pm$ 0.30\\ 
  
  & {\color{blue}A-GCE} & 59.69 $\pm$ 0.24\scriptsize{\color{red}{$\uparrow$0.92}} & 50.37 $\pm$ 0.32\scriptsize{\color{red}{$\uparrow$1.52}} & 39.66 $\pm$ 0.32\scriptsize{\color{red}{$\uparrow$0.92}} & 30.42 $\pm$ 0.16\scriptsize{\color{red}{$\uparrow$0.83}}\\ 
  \cline{2-6}
  
  & SL & 57.84 $\pm$ 0.19 & 49.37 $\pm$ 0.08 & 38.84 $\pm$ 0.18 & 29.17 $\pm$ 0.35\\ 
  
  & {\color{blue}A-SL} & 61.11 $\pm$ 0.11\scriptsize{\color{red}{$\uparrow$3.27}} & 51.14 $\pm$ 0.24\scriptsize{\color{red}{$\uparrow$1.77}} & 39.35 $\pm$ 0.21\scriptsize{\color{red}{$\uparrow$0.51}} & \textbf{30.60} \bm{$\pm$} \textbf{0.23}\scriptsize{\color{red}{$\uparrow$1.43}}\\ 
  \cline{2-6}
  
  & PolySoft & 56.73 $\pm$ 0.10 & 48.52 $\pm$ 0.31 & 39.35 $\pm$ 0.19 & 29.64 $\pm$ 0.25\\ 
  
  & {\color{blue}A-PolySoft} & \textbf{61.49} \bm{$\pm$} \textbf{0.09}\scriptsize{\color{red}{$\uparrow$4.76}} & \textbf{53.48} \bm{$\pm$} \textbf{0.21}\scriptsize{\color{red}{$\uparrow$4.96}} & \textbf{46.57} \bm{$\pm$} \textbf{0.12}\scriptsize{\color{red}{$\uparrow$7.22}} & 30.22 $\pm$ 0.15\scriptsize{\color{red}{$\uparrow$0.58}}\\ 
  \cline{2-6}
  
  & JS & 56.57 $\pm$ 0.16 & 48.73 $\pm$ 0.35 & 38.92 $\pm$ 0.26 & 29.66 $\pm$ 0.25\\ 
  
  & {\color{blue}A-JS} & 60.27 $\pm$ 0.28\scriptsize{\color{red}{$\uparrow$3.70}} & 51.91 $\pm$ 0.17\scriptsize{\color{red}{$\uparrow$3.18}} & 39.81 $\pm$ 0.40\scriptsize{\color{red}{$\uparrow$0.89}} & 29.70 $\pm$ 0.28\scriptsize{\color{red}{$\uparrow$0.14}}\\ 
  \cline{2-6}
  
  \hline
  
  \end{tabular}}
\end{table*}

\subsection{Instance-dependent Label Noise}
\label{section52}

\textbf{Datasets.} We follow the instance-dependent label noise generation scheme presented in \cite{xia2020part}. Firstly, we use a truncated normal distribution $N(\eta,0.1^{2},[0.1])$ to sample the flip rate $q$ for sample $(\mathbf{x}_{i},\mathbf{y}_{i})$, where $[0.1]$ is the range of flip rate and $\eta$ is the global noise rate. Then, we sample a tensor $W$ with the size $s\times c$ from a standard normal distribution for every sample, where $s$ is the length of sample (we change the dimension of the feature to $1\times s$). And we acquire the flip rate for each class by calculating $\mathbf{p} = q\cdot softmax(\mathbf{x}_{i}\cdot W)$ with the $(\mathbf{x}_{i}\cdot W)_{\mathbf{y}_{i}}=-\infty$. At last, we denote the $\mathbf{p}_{\mathbf{y}_{i}}=1-q$ to make sure the sum of each components of $\mathbf{p}$ is 1 and the noisy label of the sample can be generated by applying the probability vector $\mathbf{p}$.

\textbf{Results.} Table \ref{table2} exhibits the classification accuracy of different competing methods under instance-dependent noise and we use the average results of last 5 epochs. It can be seen that the results of the proposed approach is evidently better than other competing robust loss functions with conventional hyperparameter setting rules. As shown in Table \ref{table2}: 1) Compared to other conventional robust loss, the PolySoft loss performs relatively worse when noise rate is less than 0.5 on CIFAR-10, and A-PolySoft outstandingly improves the accuracy, over 10\% under different noise rates. Besides, A-PolySoft shows the best results on CIFAR-100 under almost all noise rates. 2) When the noise rate is less than 0.4 on CIFAR-10, the performance of A-SL is the best. The performance of SL drops fast as the noise rate exceeds 0.4, but other three robust loss functions with instance-dependent hyperparameters can still maintain better results, especially A-PolySoft which improves over 10\% on CIFAR-10 with 50\% noise rate. 3) Comparing to symmetric and asymmetric label noise, instance-dependent label noise is more complex and the performance of original robust losses drops sharply. The proposed noise-aware robust losses reduce performance degradation. 4) Similar to instance-independent noise, the results of conventional robust losses on instance-dependent noise degrades significantly as the noise rate rises and the instance-dependent hyperparameters help robust losses reduce the negative effects brought by the change of noise rate. The phenomena involved in (3) and (4) verify that under the more complicated instance-dependent label noise, the noise-awareness (as shown in Fig.\ref{figure1}, samples with distinct noise properties should be assigned different hyperparameters to weaken or strengthen their influence) can also be conductive to improving the noise-tolerant ability of robust losses.

\begin{table}[!t]
  \renewcommand\arraystretch{1.2}
  \centering
  \caption{Test accuracy (\%) of competing methods on CIFAR-10, CIFAR-100 under instance-independent noise. The best results are in bold.}
  \label{table3}
  \resizebox{\linewidth}{!}{
  \begin{tabular}{c|c|c|c|c|c|c}
  \hline
  
  \multirow{2}{*}{Datasets} & \multirow{2}{*}{Methods} & \multicolumn{4}{c|}{Symmetric} & \multicolumn{1}{c}{Asymmetric}\\
  \cline{3-7}
  
  & & \multicolumn{1}{c|}{0.2} & \multicolumn{1}{c|}{0.5} & \multicolumn{1}{c|}{0.8} & \multicolumn{1}{c|}{0.9} & \multicolumn{1}{c}{0.4}\\
  \hline
  \hline
  
  \multirow{6}{*}{CIFAR-10} & DivideMix \cite{li2020dividemix} & 96.1 & 94.6 & 93.2 & 76.0 & 93.4 \\ 
  
  & ELR+ \cite{liu2020early} & 94.6 & 93.8 & 91.1 & 75.2 & 92.7 \\ 
  
  & A-GCE & 96.1 & \textbf{95.3} &	\textbf{93.7} & 88.8 &	95.1 \\
  
  & A-SL & 96.0 & 94.9 & 93.0 & \textbf{90.7} & 95.4\\

  & A-Polysoft & 96.0 &	94.7 & 93.2 &	84.0 & \textbf{95.9} \\

  & A-JS & \textbf{96.3} &	95.2 & 93.3 &	88.5 & 95.0 \\
  
  \hline

  \multirow{6}{*}{CIFAR-100} & DivideMix \cite{li2020dividemix} & 77.3 & 74.6 & 60.2 & 31.5 & 31.5 \\

  & ELR+ \cite{liu2020early} & 77.5 & 72.4 & 58.2 & 30.8 & \textbf{76.5} \\ 
  
  & A-GCE & 78.5 & \textbf{75.9} & 67.8 & 38.8 & 73.5 \\
  
  & A-SL & 78.9 & 75.8 & \textbf{69.0} & \textbf{51.0} & 74.2 \\

  & A-Polysoft & \textbf{79.6} & 75.4 & 61.6 & 32.0 & 73.6 \\

  & A-JS & 77.9 & 75.6 & 66.8 & 46.1 & 72.9 \\
  
  \hline
  
  \end{tabular}}
\end{table}

\begin{table}[!t]
  \renewcommand\arraystretch{1.2}
  \centering
  \caption{Test accuracy (\%) of competing methods on mini-WebVision. The best results are in bold.}
  \label{table4}
  \resizebox{\linewidth}{!}{
  \begin{tabular}{c|c|c|c|c}
  \hline
  
  \multirow{2}{*}{Methods} & \multicolumn{2}{c|}{WebVision} & \multicolumn{2}{c}{ILSVRC12}\\
  \cline{2-5}
  
  & \multicolumn{1}{c|}{Top1} & \multicolumn{1}{c|}{Top5} & \multicolumn{1}{c|}{Top1} & \multicolumn{1}{c}{Top5}\\
  \hline
  \hline
  
  MentorNet \cite{jiang2018mentornet} & 63.00 & 81.40 & 57.80 & 79.92 \\ 
  
  Co-teaching \cite{han2018co} & 63.58 & 85.20 & 61.48 & 84.70 \\
  
  DivideMix \cite{li2020dividemix} & 77.32 & 91.64 & 75.20 & 90.84 \\

  ELR+ \cite{liu2020early} & 77.78 & 91.68 & 70.29 & 89.76 \\

  A-GCE & 78.00 & 92.88 & 75.48 & 93.28\\

  A-SL & \textbf{78.92} & 93.12 & \textbf{75.64} & 92.68 \\

  A-Polfsoft & 78.24 & 92.72 & 75.08 & 92.76 \\

  A-JS & 77.76 & \textbf{93.48} & 75.20 & \textbf{93.92} \\
  
  \hline
  
  \end{tabular}}
\end{table}

\subsection{Compared with SOTA methods}
\label{section53}

\textbf{Datasets.} We use the same datasets as described in Section \ref{section51}, which are CIFAR-10 and CIFAR-100 datasets containing instance-independent label noise. Besides, we consider a real-world noisy dataset WebVision \cite{li2017webvision}, a large-scale image dataset collected by crawling Flickr and Google, which contains about 20\% noisy labels. Here, we adopt the mini-WebVision dataset \cite{jiang2018mentornet} consisting of the top 50 classes of the Google subset.

\textbf{Results.} The noise-aware robust loss approach exhibits superior performance than conventional robust loss methods. However, when compared to the SOTA methods for addressing noisy labels, such as DivideMix \cite{li2020dividemix}, ELR+ \cite{liu2020early}, there still exists a gap. To address this limitation, we employ mixup \cite{zhang2017mixup} and consistency regularization \cite{oliver2018realistic} techniques, which have been demonstrated to be useful in robust learning \cite{li2020dividemix}, to enhance our approach. As evidenced by the results presented in Table. \ref{table3} and Table. \ref{table4}, our method outperforms the SOTA method in both extreme synthetic noise scenarios and real-world noise.

\begin{figure}[!t]
  \centering
  \subfigure[A-GCE]{\includegraphics[width=0.14\textwidth]{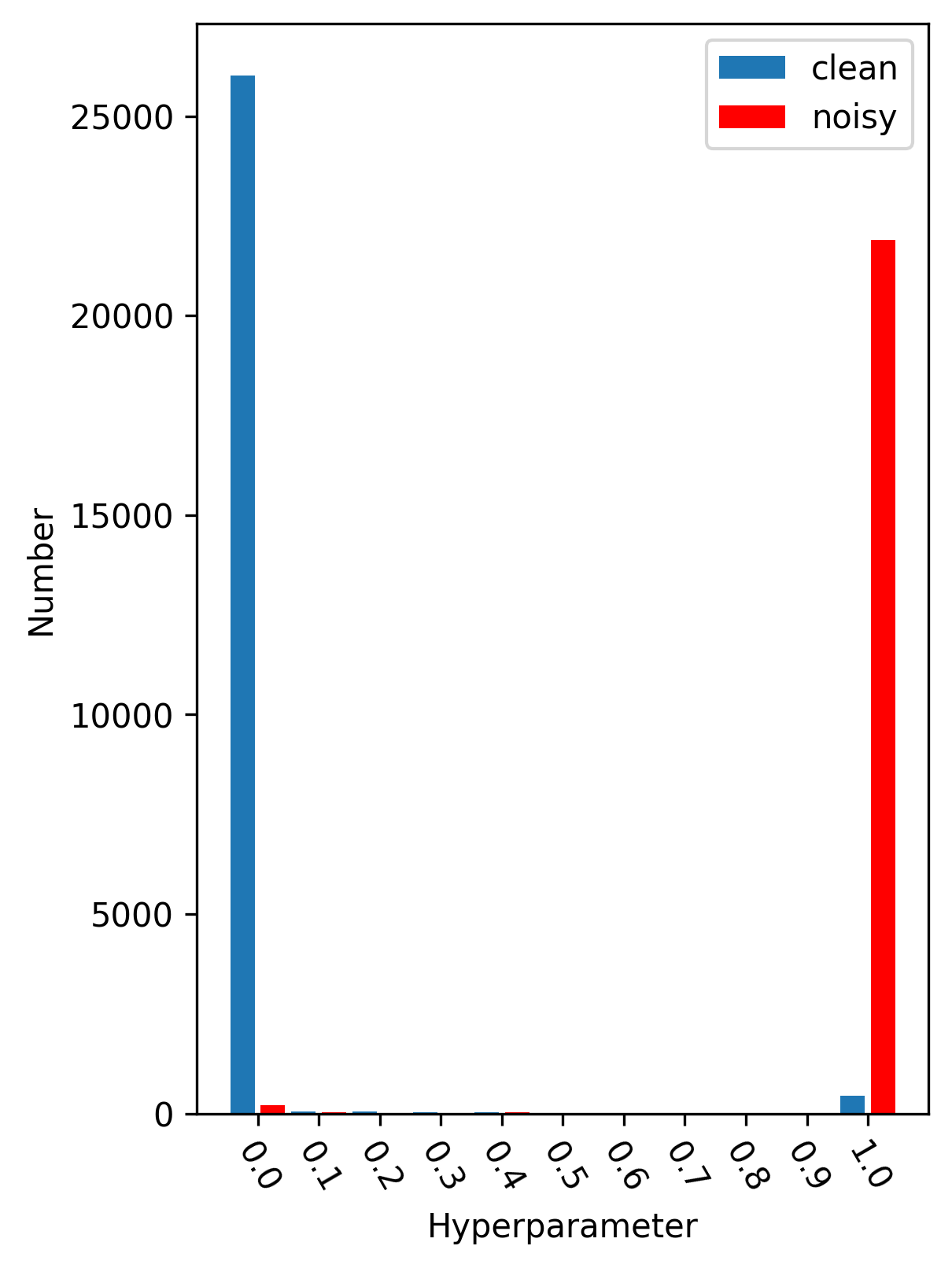}} 
  \subfigure[A-SL]{\includegraphics[width=0.14\textwidth]{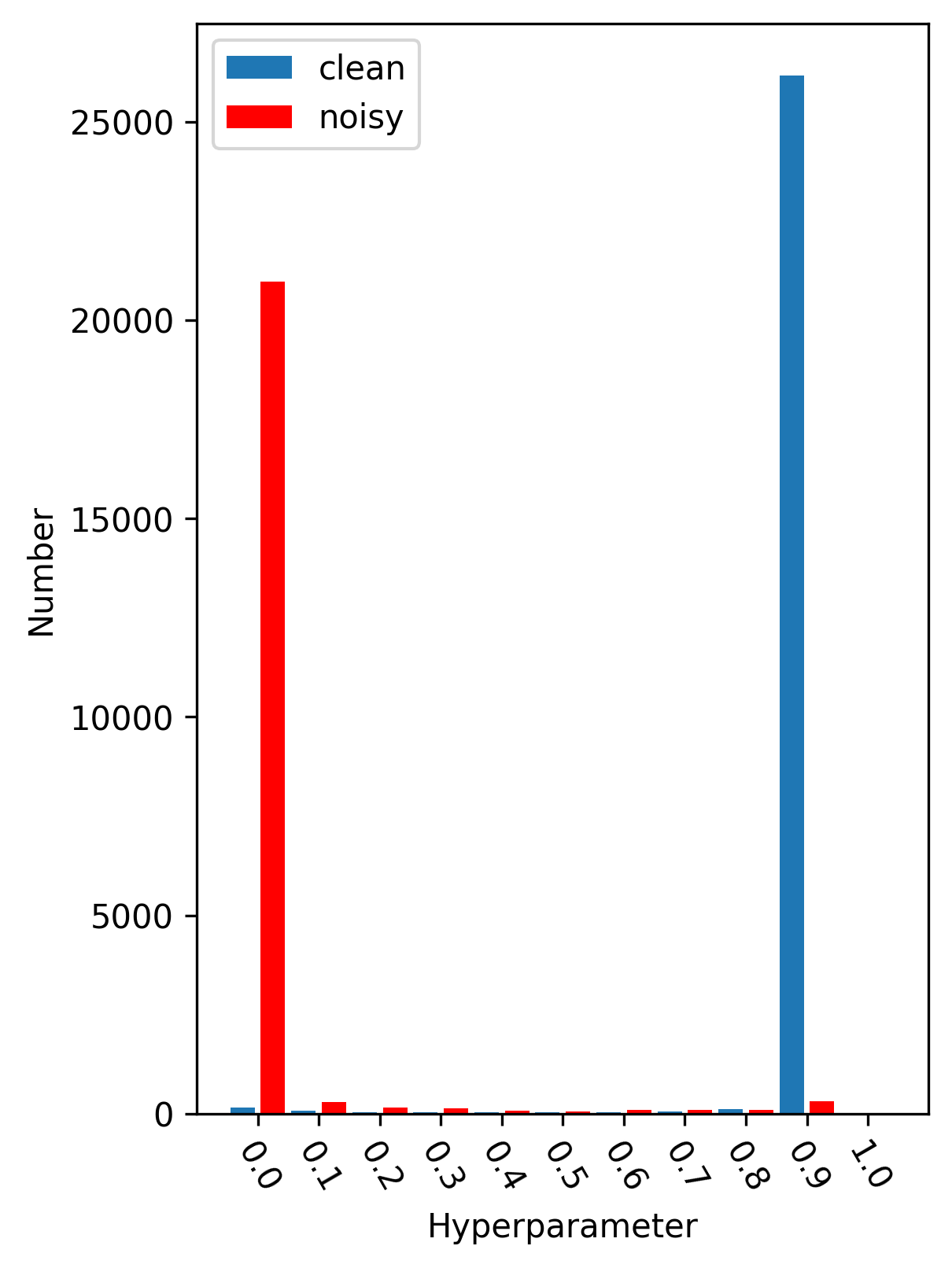}}
  \subfigure[A-JS]{\includegraphics[width=0.14\textwidth]{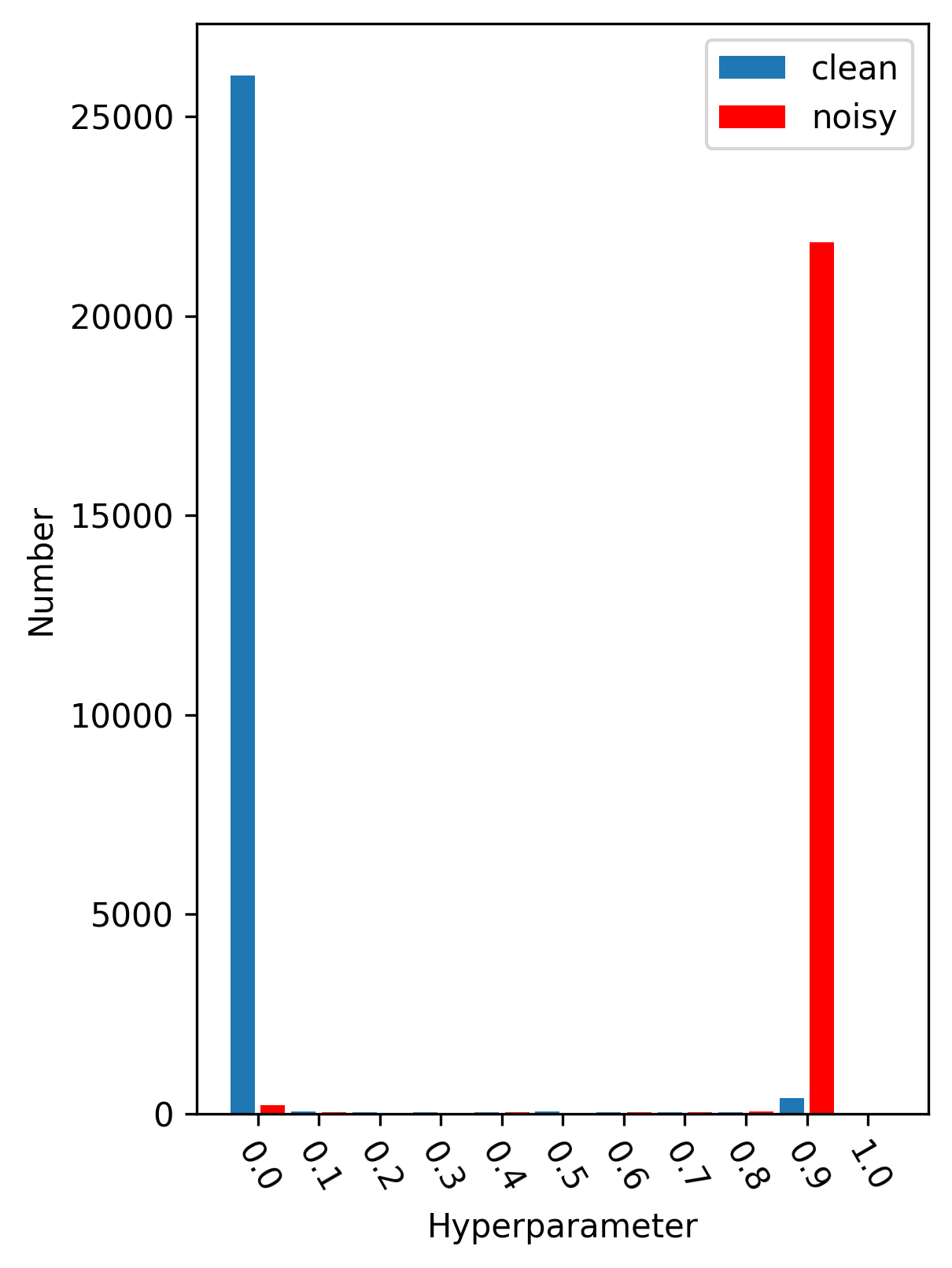}}
  \caption{Hyperparameter distribution of training samples on CIFAR-10 dataset under 50\% uniform noise experiments during meta-train process. (a)-(c) present the sample hyperparameters produced by A-GCE, A-SL and A-JS (we exhibit ${\gamma_1}$ for A-SL).}
  \label{figure6}
\end{figure}

\begin{figure}[t]
	\centering
  \setlength{\abovecaptionskip}{-0.25cm}
	\includegraphics[width=0.36\textwidth]{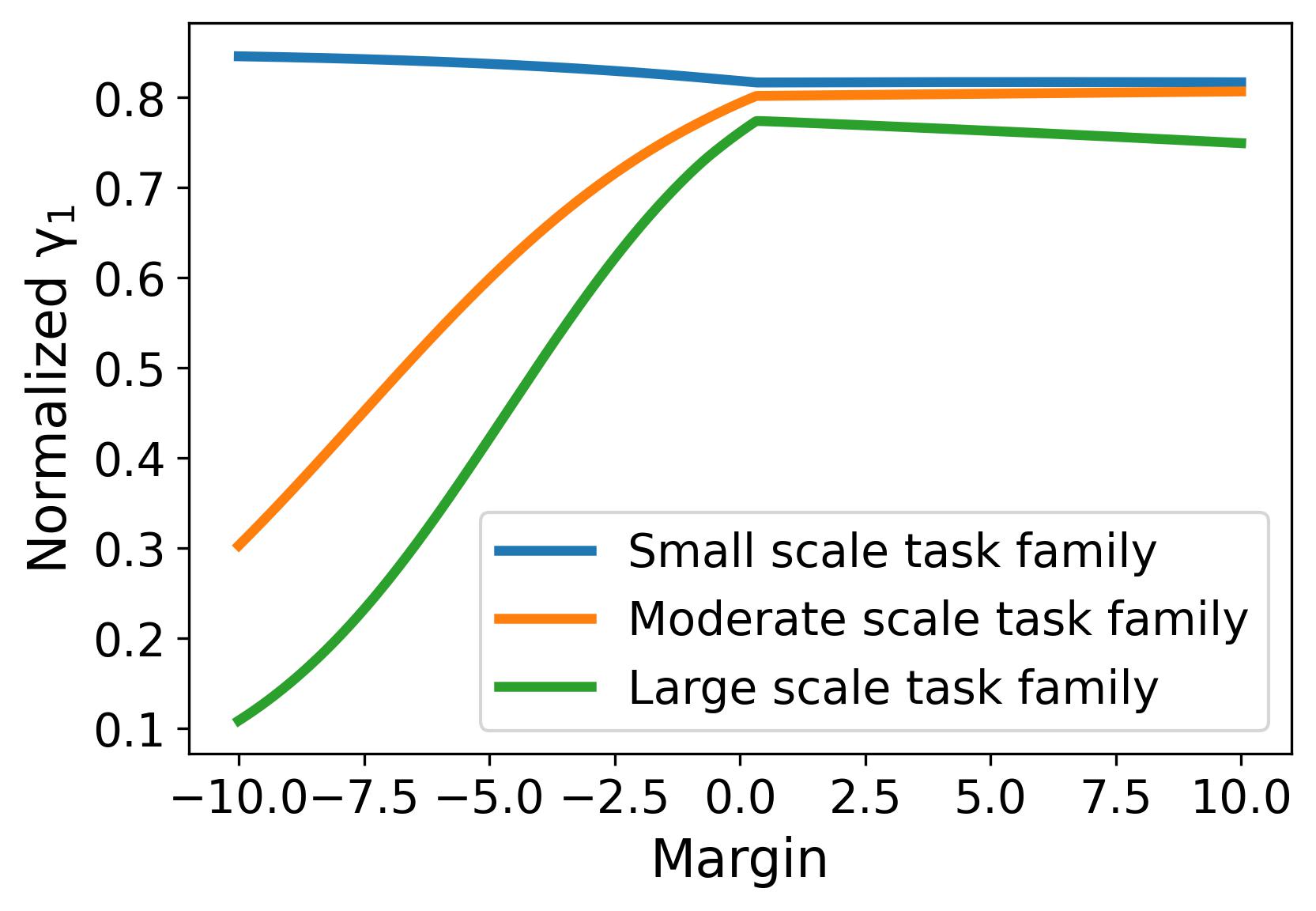}
	\caption{Hyperparameter prediction functions of SL learned by NARL-Adjuster on mini-WebVision.}
	\label{figure7}
  \vspace{-0.6cm}
\end{figure}

\subsection{Towards Understanding of NARL-Adjuster Algorithm}
\label{section54}

\textbf{The noise-awareness of robust loss.} From Pro. \ref{pro1}, it can be inferred that achieving noise-awareness for robust losses requires the assignment of distinct hyperparameters to samples with different margins. The margin information of a sample not only characterizes its influence on the decision boundary but also serves to distinguish between clean and noisy samples in the context of label noise issues. Therefore, we aspire to satisfy the following conditions for the noise-robust factors of the loss functions: for GCE and JS, the hyperparameters of clean data satisfy $q, \pi_1\rightarrow0$ and the hyperparameters of noisy data satisfy $q=1$ and $\pi_1\rightarrow1$; Additionally, considering SL as a linear combination of CE and MAE, the coefficient of CE loss on clean samples should be larger than that on noisy samples. As shown in the Fig.\ref{figure6}, the meta-learned NARL-Adjusters capture the underlying mapping relationship between instance-dependent features and hyperparameters which is consistent with the theoretical conclusions in Pro. \ref{pro1}. Most clean samples and noisy samples are properly separated and equipped with hyperparameters representing noise robustness and capability to learn sufficiently. Compared to instance-independent hyperparameter setting strategies, these hyperparameter distributions reflect that the proposed algorithm realizes strengthening the positive influence of clean samples and weakening the negative influence of noisy samples which enhances the robust loss functions to be noise-aware.

In Fig.\ref{figure7}, we exhibit the meta-learned hyperparameter prediction functions of SL. It can be observed that, due to the structure of NARL-Adjuster (Fig.\ref{figure3}), distinct hyperparameter prediction functions are acquired for categories with different noise properties. For instance, in scenarios involving categories with a smaller number of training samples, NARL-Adjuster tends to amplify the positive impact of all samples during training, i.e. the coefficient of CE in SL ($\gamma_1$ in Eq.\ref{eq3}) is relatively higher. Conversely, for the other two cases, NARL-Adjuster learns the aforementioned noise-awareness.

\begin{figure}[t]
  \centering
  \subfigure[GCE]{\includegraphics[width=0.24\textwidth]{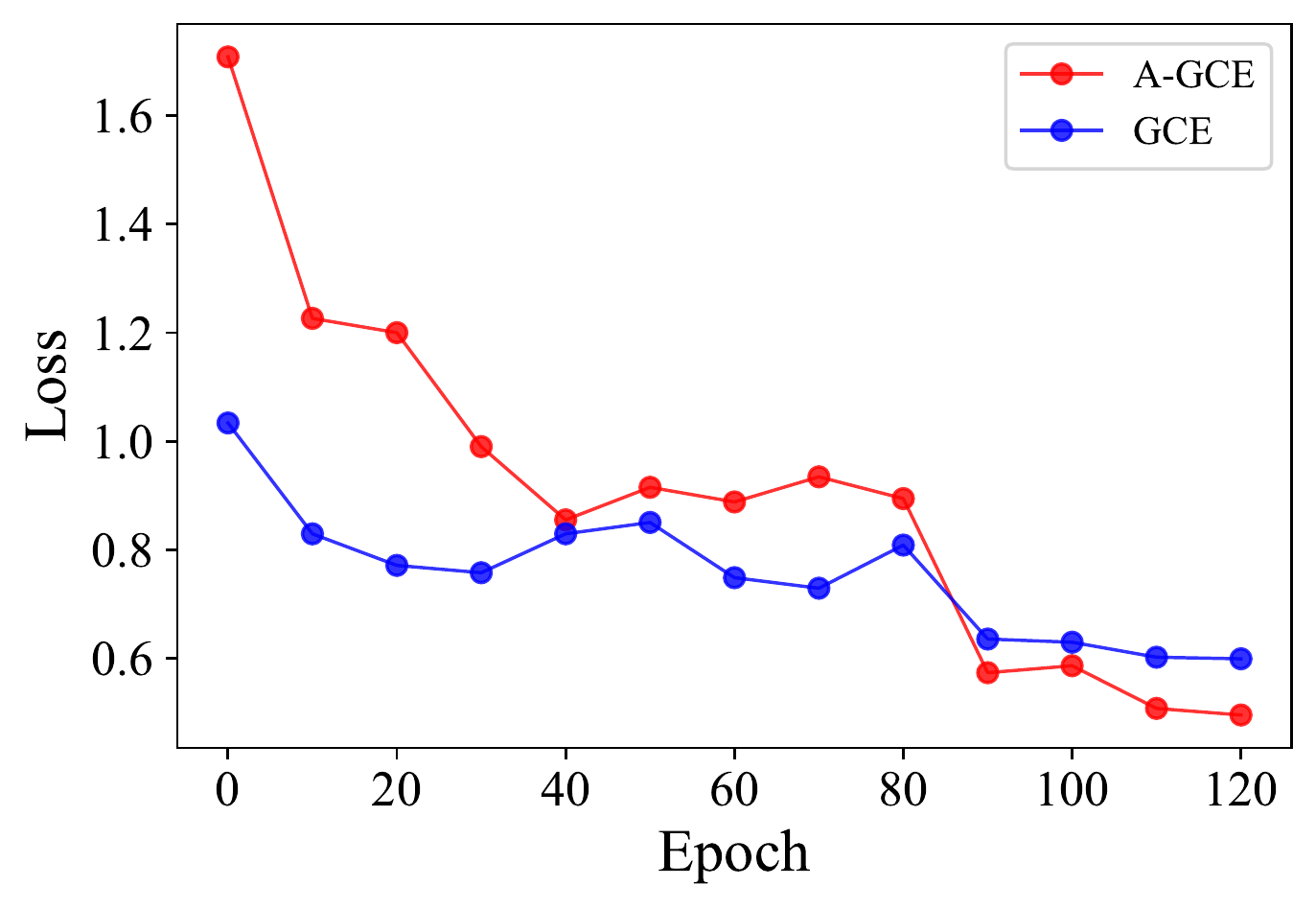}} 
  \subfigure[JS]{\includegraphics[width=0.24\textwidth]{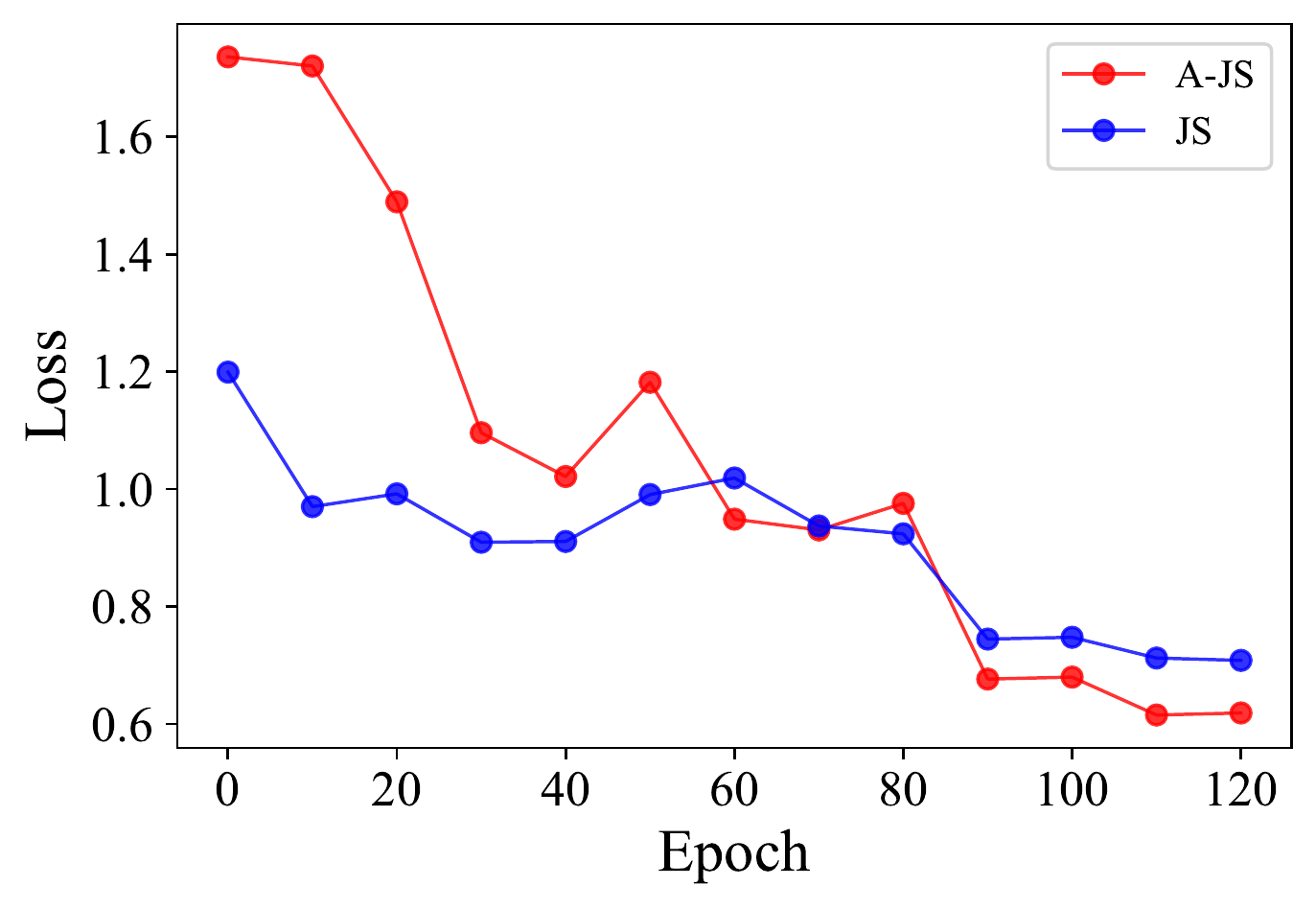}}
  \caption{The training loss vs. epoch of conventional robust loss ((a) GCE; (b) JS) and robust loss with meta-learned hyperparameter setting ((a) A-GCE; (b) A-JS) on CIAFR-10 dataset under 40\% uniform noise.}
  \label{figure8}
\end{figure}

\textbf{The training loss variation.} We have verified the validity and rationality of noise-aware hyperparameter setting rules from both theoretical (Section \ref{section34}) and experimental (Fig.\ref{figure6}) perspectives. Here, we try to verify the value of theory via observing the variation of training loss. In Fig.\ref{figure8}, with epoch increasing, the training losses of A-GCE and A-JS continue to decrease and finally lower than those of conventional robust losses which means the proposed approach benefits from enhanced noise-tolerance (as shown in Thm.\ref{thm2}), leading to find a better solution.

\begin{table*}[t]
  \renewcommand\arraystretch{1.2}
  \centering
  \caption{Test accuracy (\%) of competing methods on TinyImageNet under different noise rates. The best results are in bold.}
  \label{table5}
  \resizebox{\linewidth}{!}{
  \begin{tabular}{c|c|l|l|l|l|l|l}
  \hline
  
  \multirow{2}{*}{Dataset} & \multirow{2}{*}{Methods} & \multicolumn{4}{c|}{Symmetric Noise} & \multicolumn{2}{c}{Asymmetric Noise}\\
  \cline{3-8}
  
  & & \multicolumn{1}{c|}{0} & \multicolumn{1}{c|}{0.2} & \multicolumn{1}{c|}{0.4} & \multicolumn{1}{c|}{0.6} & \multicolumn{1}{c|}{0.2} & \multicolumn{1}{c}{0.4}\\
  \hline
  \hline
  
  \multirow{11}{*}{T-ImageNet} & CE & 55.69 $\pm$ 0.09 & 44.20 $\pm$ 0.11 & 33.95 $\pm$ 0.26 & 24.15 $\pm$ 0.42 & 46.18 $\pm$ 0.25 & 34.12 $\pm$ 0.09 \\ 
  
  & Forward & 55.90 $\pm$ 0.14 & 44.68 $\pm$ 0.16 & 36.47 $\pm$ 0.23 & 27.65 $\pm$ 0.42 & 47.25 $\pm$ 0.31 & 35.04 $\pm$ 0.23 \\ 
  
  & MW-Net & 56.53 $\pm$ 0.50 & 49.04 $\pm$ 0.73 & 43.74 $\pm$ 0.40 & 34.75 $\pm$ 0.17 & 48.20 $\pm$ 0.55 & 35.11 $\pm$ 0.35 \\ 
  \cline{2-8}
  
  & GCE & 54.07 $\pm$ 0.25 & 48.26 $\pm$ 0.22 & 43.75 $\pm$ 0.14 & 35.39 $\pm$ 0.10 & 48.70 $\pm$ 0.12 & 34.07 $\pm$ 0.42 \\ 
  
  & {\color{blue}T-GCE} & 56.04 $\pm$ 0.15 \scriptsize{\color{red}{$\uparrow$1.97}} & 51.47 $\pm$ 0.36 \scriptsize{\color{red}{$\uparrow$3.21}} & 45.51 $\pm$ 0.38 \scriptsize{\color{red}{$\uparrow$1.76}} & 38.84 $\pm$ {0.29} \scriptsize{\color{red}{$\uparrow$3.45}} & 50.10 $\pm$ 0.26 \scriptsize{\color{red}{$\uparrow$1.40}} & \textbf{36.07} $\pm$ \textbf{0.26} \scriptsize{\color{red}{$\uparrow$2.00}} \\ 
  \cline{2-8}
  
  & SL & 56.98 $\pm$ 0.17 & 48.97 $\pm$ 0.32 & 41.90 $\pm$ 0.34 & 29.97 $\pm$ 0.31 & 45.41 $\pm$ 0.20 & 33.29 $\pm$ 0.23 \\ 
  
  & {\color{blue}T-SL} & \textbf{57.30} \bm{$\pm$} \textbf{0.22} \scriptsize{\color{red}{$\uparrow$0.32}} & 52.24 $\pm$ 0.13 \scriptsize{\color{red}{$\uparrow$3.27}} & {45.73} $\pm$ {0.24} \scriptsize{\color{red}{$\uparrow$3.83}} & 35.46 $\pm$ 0.19 \scriptsize{\color{red}{$\uparrow$5.49}} & 50.30 $\pm$ 0.15 \scriptsize{\color{red}{$\uparrow$4.89}} & 34.32 $\pm$ 0.21 \scriptsize{\color{red}{$\uparrow$1.03}} \\
  \cline{2-8}
  
  & PolySoft & 55.98 $\pm$ 0.16 & 51.35 $\pm$ 0.16 & 41.77 $\pm$ 0.15 & 29.70 $\pm$ 0.53 & 46.21 $\pm$ 0.25 & 33.99 $\pm$ 0.19 \\ 
  
  & {\color{blue}T-PolySoft} & 56.47 $\pm$ 0.16 \scriptsize{\color{red}{$\uparrow$0.49}} & \textbf{52.64} \bm{$\pm$} \textbf{0.20} \scriptsize{\color{red}{$\uparrow$1.29}} & 45.38 $\pm$ 0.21 \scriptsize{\color{red}{$\uparrow$3.61}} & 35.43 $\pm$ 0.23 \scriptsize{\color{red}{$\uparrow$5.73}} & 50.82 $\pm$ 0.26 \scriptsize{\color{red}{$\uparrow$4.61}} & 34.61 $\pm$ 0.16 \scriptsize{\color{red}{$\uparrow$0.62}} \\
  \cline{2-8}
  
  & JS & 54.67 $\pm$ 0.24 & 46.26 $\pm$ 0.10 & 42.08 $\pm$ 0.09 & 34.78 $\pm$ 0.17 & 46.72 $\pm$ 0.27 & 34.40 $\pm$ 0.27\\ 
  
  & {\color{blue}T-JS} & 54.67 $\pm$ 0.21 & 51.78 $\pm$ 0.17 \scriptsize{\color{red}{$\uparrow$5.52}} & \textbf{48.34} \bm{$\pm$} \textbf{0.19} \scriptsize{\color{red}{$\uparrow$6.26}} & \textbf{40.72} \bm{$\pm$} \textbf{0.15} \scriptsize{\color{red}{$\uparrow$5.94}} & \textbf{52.18} \bm{$\pm$} \textbf{0.12} \scriptsize{\color{red}{$\uparrow$5.46}} & 35.47 $\pm$ 0.20 \scriptsize{\color{red}{$\uparrow$1.07}} \\ 
  
  \hline
  
  \end{tabular}}
\end{table*}

\subsection{Meta-Test: Deploying our meta-learned NARL-Adjuster on unseen datasets }
\label{section55}

As alluded to above, our NARL-Adjuster with an explicit parameterized structure is able to learn the common hyperparameter setting rule for various noisy label problems. This fine property implies that our meta-learned NARL-Adjuster is potentially transferrable, and can be directly deployed on unseen noisy label tasks in a plug-and-play way. To this goal, we attempt to learn a suitable NARL-Adjuster from relatively small-scale datasets, for example, CIFAR-10, and use meta-learned NARL-Adjuster as a off-the-shelf robust loss setting regime for larger datasets or real datasets. Specifically speaking, in order to alleviate meta-learned NARL-Adjuster overfitting on specific noise rate or noise type, we adopt a muti-task framework followed current meta-learning setting \cite{hospedales2021meta,JMLR:v24:21-0742} containing 9 datasets (CIFAR-10) with three noise types under three noise rates (20\%, 40\% and 60\%) in meta-train stage. The experimental setup is the same as Section \ref{section51} except the learning rate decays 0.1 at 40 epochs and 50 epochs for a total of 60 epochs. We then deploy meta-learned NARL-Adjuster on new classification tasks with noisy labels to validate its transferability.

\begin{table}[t]
  \renewcommand\arraystretch{1.2}
  \centering
  \caption{Test accuracy (\%) of competing methods on real-world noisy datasets. The best results are in bold.}
  \label{table6}
  \begin{tabular}{c|l|l}
  \hline
  
  \multirow{2}{*}{Methods} & \multicolumn{2}{c}{Datasets}\\
  \cline{2-3}
  
  & \multicolumn{1}{c|}{ANIMAL-10N} & \multicolumn{1}{c}{Food-101N} \\
  
  \hline
  \hline
  
  CE & 79.40 $\pm$ 0.14 & 81.44 $\pm$ 0.10\\ 
  
  Forward & 80.91 $\pm$ 0.05 & 84.22 $\pm$ 0.09\\
  \hline
  
  GCE & 81.05 $\pm$ 0.13 & 84.52 $\pm$ 0.15\\ 
  
  {\color{blue}T-GCE} & 82.22 $\pm$ 0.08\scriptsize{\color{red}{$\uparrow$1.17}} & 85.47 $\pm$ 0.11\scriptsize{\color{red}{$\uparrow$0.95}}\\ 
  \hline
  
  SL & 79.78 $\pm$ 0.24 & 83.93 $\pm$ 0.10\\
  
  {\color{blue}T-SL} & 81.35 $\pm$ 0.16\scriptsize{\color{red}{$\uparrow$1.57}} & \textbf{85.90} $\pm$ \textbf{0.01}\scriptsize{\color{red}{$\uparrow$1.97}}\\
  \hline
  
  PolySoft & 80.90 $\pm$ 0.27 & 84.70 $\pm$ 0.08\\ 
  
  {\color{blue}T-PolySoft} & \textbf{82.90} $\pm$ \textbf{0.21}\scriptsize{\color{red}{$\uparrow$2.00}} & 85.33 $\pm$ 0.06\scriptsize{\color{red}{$\uparrow$0.63}}\\
  \hline
  
  JS & 81.11 $\pm$ 0.20 & 82.05 $\pm$ 0.12\\
  
  {\color{blue}T-JS} & 82.86 $\pm$ 0.11\scriptsize{\color{red}{$\uparrow$1.75}} & 85.38 $\pm$ 0.15\scriptsize{\color{red}{$\uparrow$3.33}}\\
  \hline
  
  \end{tabular}
\end{table}

\textbf{Datasets.} We firstly verify the transferability of meta-learned NARL-Adjuster on a larger and harder dataset called TinyImageNet (T-ImageNet briefly), containing 200 classes with 100K training, 10K validation, 10K test images of 64 × 64. Besides, we test on three real-world noisy datasets: ANIMAL-10N \cite{song2019selfie}, consisting of 10 animals which can be divided into 5 pairs with confusing appearance. The datasets has 50,000 training images with about 8\% label noise and 5,000 testing images. Food-101N \cite{lee2018cleannet} is a dataset containing about 310k images of food recipes classified in 101 categories. The 25K test images with curated annotations are from Food-101 \cite{bossard2014food} and the noise rate is around 20\%.

\textbf{Results.}
We summarize the test accuracy on T-ImageNet with different noise settings and real-world noisy datasets in Table \ref{table5} and Table \ref{table6} (T-GCE, T-SL and T-Polysoft represent the robust losses with transferrable hyperparameter prediction functions NARL-Adjuster). We report the average results of last 5 epochs. 
It can be found that our transferrable NARL-Adjuster can consistently improve the performance of the original robust loss methods on these tasks. For example, the results of SL and PolySoft drop quickly when the noise rate exceeds 0.4 on T-ImageNet dataset, while T-SL and T-PolySoft can improve around 4\% points. Under symmetric label noise, The performance gains of T-GCE and T-JS increase as noise rate raises. On the contrary, under asymmetric noise cases, the performance gains decrease with noise rate reducing. These robust loss functions are more robust under symmetric noise may caused the above phenomenon.

Note that our plug-and-play NARL-Adjuster is easy to be deployed on unseen tasks with lightweight computation. Comparatively, original robust loss methods require careful hyperparameter tuning with relatively heavy computation burden. Such property is promising to deploy our method on real applications. As shown in Table \ref{table6}, the robust loss methods equipped with our transferrable NARL-Adjuster have an evident improvement in noise tolerance on real-world noisy datasets while with evidently less additional hyperparameter tuning process. This demonstrates that our NARL-Adjuster is hopeful to reduce the barrier of practical implementations for current robust loss methods.

\section{Conclusion}
In this paper, we have proposed to equip robust loss function with noise-aware hyperparameters to improve their noise tolerance with theoretical guarantee. We use an explicit hyperparameter prediction function, called NARL-Adjuster to predict instance-dependent hyperparameters, which can be learned from data in a meta-learning way. The proposed NARL-Adjuster is model-agnostic and has been substantiated to be valid in improving the noise-aware ability of four kinds of SOTA robust loss methods with comprehensive experiments under various noise structure forms. The meta-learned NARL-Adjuster is plug-and-play and can be deployed on unseen tasks with negligible hyperparameter tuning cost. This property makes it potential to reduce the barrier of heavy hyperparameter tuning burden for current robust loss methods on real problems.
Provable transferrable hyperparameter prediction function theory, as done for methodology-level transfer \cite{JMLR:v24:21-0742}, is interesting for future research.

\bibliographystyle{IEEEtran}

\bibliography{my.bib}

\end{document}